\documentclass[12pt]{amsart}
\usepackage{amsbsy,amssymb,amsmath,amsthm,amscd,amsfonts,latexsym,amstext,delarray,
amsmath,graphicx,url} 
\usepackage{qtree}
\usepackage[margin=.8in]{geometry}
\usepackage{color}
\usepackage[all]{xy}

\newtheorem{thm}{Theorem}[section]
\newtheorem{prop}[thm]{Proposition}
\newtheorem{cor}[thm]{Corollary}
\newtheorem{lem}[thm]{Lemma}

\newtheorem{defn}[thm]{Definition}
\newtheorem{rem}[thm]{Remark}

\numberwithin{equation}{section}

\def\F{{\mathbb F}}
\def\Q{{\mathbb Q}}
\def\Z{{\mathbb Z}}
\def\N{{\mathbb N}}

\def\cA{{\mathcal A}}
\def\cB{{\mathcal B}}
\def\cC{{\mathcal C}}

\def\cE{{\mathcal E}}

\def\cH{{\mathcal H}}
\def\cI{{\mathcal I}}

\def\cL{{\mathcal L}}

\def\cO{{\mathcal O}}
\def\cP{{\mathcal P}}

\def\cR{{\mathcal R}}
\def\cS{{\mathcal S}}

\def\cV{{\mathcal V}}
\def\cW{{\mathcal W}}

\def\fM{{\mathfrak M}}
\def\fT{{\mathfrak T}}
\def\fF{{\mathfrak F}}

\title{Mathematical Structure of Syntactic Merge}
\author{Matilde Marcolli, Noam Chomsky, Robert C.~Berwick}
\date{2023}
\address{Departments of Mathematics and of Computing and Mathematical Sciences, 
California Institute of Technology, Pasadena, CA 91125, USA}
\email{matilde@caltech.edu}
\address{Department of Linguistics, 
University of Arizona, Tucson, AZ 85721, USA}
\email{noamchomsky@email.arizona.edu}
\email{chomsky@mit.edu}
\address{Institute for Data, Systems, and Society,
Massachusetts Institute of Technology, 
Cambridge MA 02141, USA}
\email{berwick@csail.mit.edu}

\begin{document}
\maketitle

\begin{abstract}
The syntactic Merge operation of the Minimalist Program in linguistics 
can be described mathematically in terms of Hopf algebras, with a 
formalism similar to the one 
arising in the physics of renormalization. This mathematical formulation
of Merge has good descriptive power, as phenomena empirically observed
in linguistics can be justified from simple mathematical arguments.  It also
provides a possible mathematical model for externalization
and for the role of syntactic parameters. 
\end{abstract}

\section{Introduction}

Within the context of generative linguistics, the {\em Minimalist Model} was introduced
in the '90s, \cite{Chomsky95}, as a formalism that analyzes the generative
process of syntax  in terms of a basic fundamental operation, referred to as
{\em Merge}, that generates, combines, and transforms syntactic trees.
The formulation of Minimalism underwent some significant changes in recent years, after 
a simplifying reformulation, \cite{Chomsky13}, \cite{Chomsky17}, \cite{Chomsky19}, 
\cite{Chomsky21}, see also \cite{BerCh}, \cite{BerEp}, \cite{Komachi}, where the 
Merge operation is described as a combinatorial binary set formation. 

\smallskip

Our main goal here is to present a mathematical formulation of the Merge 
operator in syntax, based on Hopf algebras. 

\smallskip

The reason why Hopf algebras are the suitable mathematical setting comes from
the fact that grafting operations on trees such as Merge provide a
natural strategy for generating a hierarchy of recursively defined structures. This idea has 
been widely developed within theoretical physics, where Hopf algebras of
rooted trees and of Feynman graphs are used to analyze the combinatorics of
perturbative expansions in quantum field theory and the formalism of renormalization,
\cite{CoKr}, \cite{CoMa}, \cite{Kr}. 
The Hopf algebra formalism of perturbative quantum field theory and renormalization
has also found applications to the theory of computation, see for instance
\cite{DelMar}, \cite{Man1}, \cite{Man2}, \cite{MaPo}. 

\smallskip

In particular, an important case in physics, where recursive structures 
are built out of operators formally resembling
the syntactic Merge, is the recursive construction of solutions to the
quantum equations of motion (Dyson--Schwinger equations), \cite{Foissy}, \cite{Yeats}. 
The Dyson--Schwinger equations can be seen as a way of implementing
recursively at the quantum level the variational least action principle
characterizing the classical equations of motion (see \S 9.6 of \cite{PeS}).

\smallskip

The fact that the Hopf algebra formalism provides a natural setting for 
the formulation of recursive operations that build hierarchical structures
based on trees strongly suggests that this should also be the natural
setting for describing the properties of the Merge operators of syntax. 
We show in this paper that this is indeed the case, namely that the
same mathematical formalism that governs the recursive structures
of quantum field theory also governs Merge in the Minimalist Model.

\smallskip

The mathematical formulation of Merge that we obtain here is
not just a convenient, mathematically elegant, rephrasing of
\cite{Chomsky17}, \cite{Chomsky19},  but it has good descriptive
and predictive power. We demonstrate that by showing that
some empirically observed linguistic phenomena acquire a 
simple and direct mathematical explanation in this model.
For example, we show that some of the properties usually
required of Merge, such as not decreasing the size of
workspaces, or cancellation of deeper copies of accessible terms,
follow directly from the mathematical formalism (in particular
from the structure of the coproduct of the Hopf algebra). We
also show that a hypothetical Merge operator that instead of
binary would be $n$-ary for any $n\geq 3$ would necessarily
suffer from both undergeneration and overgeeration with
respect to the binary Merge. This property again
follows immediately from counting arguments in
the Hopf algebra and confirms and explains in
clear computational terms some empirically observed
linguistic phenomena. 

\smallskip

We also show that, within this mathematical setting, one
can formulate a possible model for the externalization
process, that interfaces the core computational mechanism
of Merge with the syntactic constraints of specific languages.
We describe the model of externalization in the form of
a correspondence, rather than a function, where one
side of the correspondence implements the linear order
of sentences, which is not present at the level of the deeper
structure of Merge, while the second step of the correspondence
implements other constraints (in addition to word order) on
the syntactic trees that come from syntactic parameters.

\smallskip

We formulate some possible questions for future study, involving
possible approaches to models of syntactic-semantic interface,
and a characterization of Merge as a solution to an optimization problem. 

\smallskip
\subsection{Summary of Merge}

We summarize briefly the structure of the Merge operation of syntax according to the 
more recent formulation of the Minimalist Model of syntax, following \cite{Chomsky17}, 
\cite{Chomsky19}. 

\smallskip

One considers a given set of {\em lexical items} and {\em syntactic features} like
$N,V,A,P,C,T,D,\ldots $. One also starts with an independently 
constructed set of {\em syntactic objects} (SO) obtained recursively, by adding to the
above set all syntactic objects created by application of a basic Merge operation.
This is defined in terms of the 
{\em binary set formation} that assigns
to two arguments $\alpha$ and $\beta$ the {\em unordered} set
\begin{equation}\label{BinSetForm}
\fM(\alpha,\beta) :=\{ \alpha, \beta \} \, ,
\end{equation}
so as to include, for 
example, sets like $\{ N, V \}$, etc. We denote by $\cS\cO$ the set obtained in
this way, starting from lexical items and syntactic features. We write $\cS\cO_0$
for the initial set of  lexical items and syntactic features. 

\smallskip

The {\em accessible terms} of a syntactic object $SO$ are proper nonempty subsets of $SO$.
(A definition in terms of binary rooted trees is given below, in Definition~\ref{acctermsDef}.)
We write $Acc(SO)$ for the set of accessible terms of a syntactic object $SO$.
A {\em Work Space} $WS$ is a finite (multi)set of syntactic objects in $\cS\cO$. 
The size of the workspace $WS$ is the sum of the number of syntactic objects 
and the number of accessible terms,
where the set $Acc(WS)$ of accessible terms of the workspace is 
$$ Acc(WS):=\bigcup_{SO\in WS} Acc(SO) \, . $$

\smallskip

Merge acting on workspaces consists of a collection of operations 
$\fM=\{ \fM_A \}$, parameterized by sets $A$ consisting
of two syntactic objects $\alpha,\beta$. These operations have as input
a workspace and produce as output a new workspace, by searching for
accessible terms in the given workspace matching the selected objects
$\alpha,\beta$, producing a new object in the workspace obtained by
applying binary set formation, and cancelling the remaining deeper
copies of the accessible terms used. 

\smallskip

The Merge action on workspaces can be given an axiomatic
formulation by imposing a list of desired properties. 
Some of the fundamental required properties of Merge are:
\begin{enumerate}
\item it is a binary operation (it applies to only two arguments in $WS$);
\item any generated syntactic object remains
accessible for further applications of Merge;
\item every accessible term only appears once in
the workspace;\footnote{It is important to distinguish here between {\em copies} and {\em repetitions}:
the workspace is a multiset of syntactic objects, hence repetitions are allowed, while this property 
refers to copies. This will be made more precise in \S \ref{WSsec} and \S \ref{CopiesSec} below.}
\item the result of Merge applied to two arguments $\alpha,\beta$ does not add any
new syntactic properties to $\alpha$ and $\beta$ nor it removes any of their existing
properties (structure preserving principle);\footnote{One should interpret this principle in the
sense that, for example, we cannot add features, or transform $\alpha$, $\beta$ into
new $\alpha’$, $\beta’$. However, syntactic properties do change through Merge, as the
following simple example suggested to us by Sandiway Fong shows: in $YP=\{ R, XP \}$
we have that $YP$ is a theta-configuration, and $XP$ has now acquired the syntactic 
property that it is the theta-marked object of $R$.}
\item workspace size does not decrease and increases at most by one.  
\end{enumerate}

\smallskip

The last condition on the size of workspaces can be broken down into two
separate conditions on the number of syntactic objects in the workspace 
and on the number of accessible terms.
The number of syntactic objects is expected to be non-increasing, and
overall decreasing in the course of the derivation, for
derivations to terminate and ``converging" thought to be generated,
while at the same time the overall number of accessible terms will be
non-decreasing, and overall increasing in 
the course of a derivation. (The term derivation is meant here in the same
sense as in logic and theory of computation).  
The condition stated as above in terms of size of the workspace is
consistent with, for example, \cite{FBG}. Separating conditions on
number of syntactic objects and of accessible terms, we can formulate
the last condition in the different form
\begin{enumerate}
\item[(5')] the number of syntactic objects in the workspace
is non-increasing, and the number of accessible terms is non-decreasing.
\end{enumerate}
We will give a more precise definition of syntactic objects, accessible terms,
and workspace size in \S \ref{SOsec} and \S \ref{WSsec}, and we
will discuss more in detail conditions (5) and (5') in \S \ref{SizeSec}. 

\smallskip

For the underlying linguistic justification for the desirable properties 
of the action of Merge on workspaces we refer the reader to 
\cite{Chomsky17}, \cite{Chomsky19}, \cite{Chomsky23}, 
and the forthcoming \cite{CSBFHKMS}. For our purposes
here, we take this list as an assigned guideline, a kind of ``axiomatic
template" on how to construct a mathematical model for the
action of Merge on workspaces. Some caveats and some more
specific interpretation of the items listed above will be discussed
in the following sections.

\smallskip

A goal of this paper is to show that such a list of fundamental properties 
leads naturally to a mathematical formulation of Merge in terms of
magmas and Hopf algebras, and that this formulation turns out to be in fact
the same very basic mathematical structure that arises naturally in the
description of fundamental interaction in physics.

\section{A mathematical model of syntactic Merge}

We consider here the formulation of Minimalism as
presented in \cite{Chomsky17}, \cite{Chomsky19}, 
with a fundamental Merge operation of binary set formation.

\subsection{Syntactic objects and the Merge magma}\label{SOsec}

As in \cite{Chomsky17}, \cite{Chomsky19}, one considers,
as the starting point in the construction of the set of
{\em syntactic objects} $\cS\cO$, an initial set, which we
denote by $\cS\cO_0$ consisting of lexical items and 
syntactic features. 

\smallskip

\begin{defn}\label{SOdef}
The set $\cS\cO$ of syntactic objects is the free, non-associative, commutative magma over 
the set $\cS\cO_0$,
\begin{equation}\label{SOeq}
\cS\cO={\rm Magma}_{na,c}(\cS\cO_0, \fM) \, ,
\end{equation}
with the binary Merge operation 
\begin{equation}\label{SOMeq}
\fM(\alpha,\beta)=\{ \alpha, \beta \} \, .
\end{equation}
\end{defn}

\smallskip

This means that, as described in \cite{Chomsky17}, \cite{Chomsky19}, the set $\cS\cO$
is obtained from the initial set $\cS\cO_0$ through iterations of the Merge operation \eqref{SOMeq}.
This procedure generates elements of $\cS\cO$ of the form $\{ \alpha, \beta \}$, 
$\{ \alpha, \{ \beta, \gamma \} \}$, for $\alpha,\beta,\gamma \in \cS\cO_0$, and so on.
The Merge operation \eqref{SOMeq} acts on the set $\cS\cO$, giving it the
structure of non-associative, commutative magma\footnote{The presence of magma structures
in generative linguistics was also recently observed independently in \cite{Coop}, in a somewhat different context.}. 

\smallskip

\begin{rem}\label{remSOtrees} {\rm
The description of the set $\cS\cO$ of syntactic objects given in Definition~\ref{SOdef}
above gives an identification 
\begin{equation}\label{SOtrees}
\cS\cO  \simeq \fT_{\cS\cO_0}
\end{equation}
of the set of syntactic objects with the set of binary, non-planar, rooted trees, with leaves labelled 
by elements of $\cS\cO_0$. } \end{rem}

\smallskip

By non-planar we mean that we regard trees $T\in \fT_{\cS\cO_0}$ as abstract trees, without
fixing a choice of a planar embedding. This implies that there is no choice of a linear ordering on
the leaves of the trees. As in \cite{Chomsky17}, \cite{Chomsky19}, the word order structure,
that is, the linearly ordered form of sentences, is considered a part of the externalization process,
not of the core computational mechanism of syntax given by Merge. 

\smallskip

\subsubsection{Planarity and lists versus sets}\label{TreesSetsSec}

In the formulation above, in Definition~\ref{SOdef} and Remark~\ref{remSOtrees}
we identify syntactic objects with {\em non-planar} binary rooted finite trees with
leaves labelled by the set $\cS\cO_0$, where
non-planar means that no choice of a planar embedding is taken for the tree.
These are often also referred to as ``abstract trees".
This is the usual mathematical description of the elements of the
free, non-associative, commutative magma on a given set.

\smallskip

While planar trees (trees together with a choice of a planar embedding) and
abstract trees appear to be similar mathematical objects, their combinatorial
properties are very different, and this accounts for several significant differences,
in linguistics, between older forms of Minimalism and the newer form we
discuss in this paper. This is discussed more explicitly in our companion
paper \cite{MBC}. 

\smallskip

In the linguistics literature, the passage from planar trees in the older
versions of Minimalism to abstract trees, is usually discussed using
the terminology {\em sets} to refer to the abstract trees as elements of
the free, non-associative, commutative magma. The reason for the use
of this terminology is that in dropping the planar structure one replaces
an identification of the set of leaves with parenthesized {\em lists} 
({\em ordered sets}, often referred to in the linguistics literature as ``strings") 
with just sets (in fact more precisely {\em multisets}). In order to avoid the 
conflict of terminology between sets and mutisets, we prefer here to
follow the standard mathematical terminology and refer to the syntactic
objects as abstract binary rooted trees (with no assigned planar embedding).

\smallskip

It is important to note that, because all the trees are binary, the clash of terminology between
sets and multisets is very mild when one considers syntactic
objects. Indeed, since trees are binary rather than $n$-ary
with some $n\geq 3$, the only repetitions of labels that give rise to 
multisets can be on two consecutive ones, so there is an unambiguous way of
labeling the same objects by sets. For example, a multiset of the form
$\{ \{ a, a \}, b, \{ c, d \} \}$ can be written equivalently as
the set $\{ \{ a \}, b, \{ c, d \} \}$ with the convention that
a set of the form $\{ a \}$ stands for the abstract tree $\Tree [ a a ]$.

\smallskip

However, even with binary trees, the clash of terminology
between sets and multisets becomes seriously problematic
when it comes to describing {\em workspaces}, as we
will see in \S \ref{WSsec} below. These are genuinely
multisets that do not have an equivalent description as sets, 
hence the mathematically correct notion to use is {\em binary forests} 
(disjoint unions of a finite collection of abstract binary rooted trees) 
rather than sets. Indeed, forests are multisets where the same tree (the same
syntactic object) may appear more than once. This is expected as 
the same syntactic object may be used repeatedly, in different ways, 
in the course of a derivation. For this reason, we will not
be using the ``sets" terminology that is more common
in the linguistics literature and we prefer to adopt the mathematical notation of
trees (with no planar structure) and forests.

\smallskip
\subsection{Workspaces: product and coproduct}\label{WSsec}

We next introduce workspaces, as in \cite{Chomsky17}, \cite{Chomsky19} and the
action of Merge on workspaces.  We first introduce workspaces with a bialgebra
structure related to the combination of workspaces and the extraction of accessible
terms with cancellation of copies.

\smallskip

\begin{defn}\label{WSdef}
Workspaces are 
nonempty finite sets of syntactic objects. The identification \eqref{SOtrees}
between syntactic objects and binary, non-planar, rooted trees,
with leaves labelled by elements of $\cS\cO_0$, induces an identification
\begin{equation}\label{WSforests}
\cW\cS \simeq \fF_{\cS\cO_0}
\end{equation}
between the set $\cW\cS$ of all workspaces and the set $\fF_{\cS\cO_0}$
of binary non-planar forests (disjoint unions of binary, non-planar, rooted trees)
with leaf labels $\cS\cO_0$. 
\end{defn}

\smallskip

Note that, with this definition of workspaces, we allow for the presence of
repeated copies of the same syntactic object in a workspace, since a forest
can have multiple connected components that are isomorphic to the same
tree. This is needed for the operations of combination of workspaces and
extraction of accessible terms described below to be well defined, as the
result of these operation can produce repeated copies of the same tree,
even when starting with a forest that has none. 

\smallskip

\begin{defn}\label{acctermsDef}
Given a binary non-planar rooted tree $T\in \fT_{\cS\cO_0}$, let $V_{int}(T)$
denote the set of all internal (non-root) vertices of $T$. 
For $v\in V_{int}(T)$,
let $T_v\subset T$ denote the subtree consisting of $v$ and all its descendants. Let $L_v=L(T_v)$
be the set of leaves of $T_v$. The set of accessible terms of $T$ is given by
\begin{equation}\label{AccTeq}
Acc(T) = \{ L_v =L(T_v) \,|\, v \in V_{int}(T) \} \, . 
\end{equation}
For a workspace given by a forest $F=\sqcup_a T_a\in \fF_{\cS\cO_0}$, the set of accessible
terms is
\begin{equation}\label{AccFeq}
Acc(F)=\bigcup_a Acc(T_a) \, , 
\end{equation}
so that we have the total number of vertices of the forest given by the sum
\begin{equation}\label{VT}
\# V(F)= b_0(F) +\# Acc(F) \, ,
\end{equation}
where $b_0(F)$ is the number of connected components (trees) of the forest $F$. 
We define the size of a workspace $F$ by
\begin{equation}\label{sizeWS}
\sigma(F) :=   \# V(F)= b_0(F) +\# Acc(F)\, ,
\end{equation}
namely the number of syntactic objects plus the total number of accessible terms.
We also define another counting function, which is given by
\begin{equation}\label{sizeWSplus}
\hat\sigma(F) := b_0(F)+  \# V(F) \, .
\end{equation}
\end{defn}

\smallskip

The size $\sigma(F)$ of the workspace, defined as in \eqref{sizeWS} is consistent with
\cite{Chomsky17}, \cite{Chomsky19} and agrees with the definition of size used in \cite{FBG}.
As pointed out to us by Riny Huijbregts, it may be preferable to consider the effect of
Merge on workspaces in terms of the counting of accessible terms $\# Acc(F)=\# V_{int}(F)$,
rather than in terms of the size $\sigma(F)$. We will discuss and compare the effect on various
size-counting in \S \ref{SizeSec} below. 

\smallskip

The set of workspaces is endowed with two operations. A product operation that combines workspaces
by taking their union is simply given by the disjoint union on the set of forests. It is a commutative
and associative product, with unit given by the empty forest. The second operation is a 
coproduct, which provides all the possible extractions of admissible terms. In order to
be able to consider all accessible terms simultaneously, one considers, instead of the set $\fF_{\cS\cO_0}$,
as above, a space of formal linear combinations of elements in this set.  Namely, we denote by
$\cV(\fF_{\cS\cO_0})$ the $\Z$-module freely generated by elements of $\fF_{\cS\cO_0}$
(formal linear combinations of binary non-planar forests with integer coefficients), so that one
can sum over all the possible extractions of accessible terms (see \eqref{coprodT} below). 

\smallskip

In the following, for a given rooted binary tree (with no assigned planar embedding) $T\in \fT_{\cS\cO_0}$,
we write $T_v$ for the subtree consisting of $v$ and all its descendants, as in Definition~\ref{acctermsDef}.
In addition to considering these sub-objects $T_v\subset T$, we also consider corresponding
quotient objects $T/T_v$. Linguistically, the $T_v$ are the accessible terms, as described above,
while the $T/T_v$ implement the cancellation of the deeper copy of $T_v$ from the resulting
workspace, after application of Merge. Mathematically, as we discuss below, the pairs $T_v$ and $T/T_v$
correspond to the two terms of a coproduct applied to $T$. The quotient object $T/T_v$ is no longer a
sub-object of $T$. 

\smallskip

The usual way of defining the quotient $T/T_v$, 
in the context of Hopf algebras of rooted trees in mathematics
and theoretical physics, is to contract the entire tree $T_v$ to a single vertex, so that the
root vertex of $T_v$ becomes a leaf of $T/T_v$. With this definition of the quotient, in
particular, one has $T/T=\bullet$, the tree consisting of a single root vertex. However,
this choice of how to define $T/T_v$ is not the best one in our setting. One reason
is that, with this definition, the new leaf of $T/T_v$ (the root vertex of $T_v$) needs to
be labeled by an element of $\cS\cO_0$, hence some calculus of labels of internal
vertices is required. While such a projection mechanism for labeling internal vertices is 
required in other versions of Minimalism, in the version we are considering we
can dispense with that, and labeling of internal vertices should not be required. 
The other reason, as we will
discuss more in detail below, is that for mathematical consistency and for a 
simple unifying view of Internal and External Merge (see Proposition~\ref{intMergecase}
below), it is necessary that the quotient $T/T=1$ is also the unit $1$ of the magma 
$\cS\cO$, and such a unit $1$
is provided not by the single vertex tree but by formally adding an empty tree.
Thus, the consistent way of defining the quotient object $T/T_v$ is provided by the
following definition.

\begin{defn}\label{Tquot}
Given a rooted binary tree (with no assigned planar embedding) $T\in \fT_{\cS\cO_0}$
and a subtree $T_v\subset T$ consisting of a vertex $v$ of $T$ and all its descendants,
the quotient $T/T_v$ is the rooted binary tree obtained by removing the entire tree $T_v$
from $T$. There is then a unique maximal rooted binary tree that can be obtained from 
the complement $T\smallsetminus T_v$ via contraction of edges. 
That resulting rooted binary tree is what we call $T/T_v$.
\end{defn}

\begin{lem}\label{Fquot}
Given a rooted binary tree (with no assigned planar embedding) $T\in \fT_{\cS\cO_0}$,
a subforest $F_{\underline{v}}\subset T$ is a union of subtrees $T_{v_1}\cup \cdots \cup T_{v_k}$,
for $\underline{v}=(v_1,\ldots, v_k)$, with the property that $T_{v_i}\cap T_{v_j}=\emptyset$
for all $i\neq j$. The quotient $T/F_{\underline{v}}$ given by
$$ T/F_{\underline{v}} = (\cdots (T/T_{v_1})/T_{v_2}\cdots )/T_{v_k}\, , $$
with each quotient of trees as in Definition~\ref{Tquot}, 
is well defined and independent of the order of $v_1,\ldots, v_k$. This extends to
quotients of the form $F/F_{\underline{v}}$ where $F=\sqcup_a T_a$ is a forest
with each component $T_a$ a rooted binary tree (with no assigned planar embedding) 
and $F_{\underline{v}}$ is a subforest of $F$ with $F_{\underline{v}}\cap T_a$ a subforest
in the sense above. In particular, these quotients of forests satisfy $F_{\underline{v},\underline{w}} /F_{\underline{v}}=F_{\underline{w}}$. 
\end{lem}

\proof  If no pair $T_{v_i}, T_{v_j}$ has a common vertex $v_{ij}$ of $T$ adjacent to both roots,
then it is clear that $(T/T_{v_i})/T_{v_j}=(T/T_{v_j})/T_{v_i}$. If a pair has such common vertex
then we can still see that this works since in this case we have
$$ T = \Tree [[  [ T_i  T_j ]  T_a  ]  T_b ] \ \ \ \ \  T/T_i = \Tree [ [ T_j  T_a ]  T_b ]  
 \ \ \ \ \  T/T_j = \Tree [ [ T_i  T_a ]  T_b ]  $$ $$  (T/T_i)/T_j = (T/T_j)/T_i = \Tree [ T_a  T_b ]  \, . $$
The extension from trees to forests follows similarly by components. For
$F_{\underline{v},\underline{w}}$ a subforest of $T$ (or of a forest $F$) on the set of
vertices $(\underline{v},\underline{w})$ with the disjointness property above, we have
that $F_{\underline{v}}$ is a subforest given by a subcollection of components, each of
which is removed in  the quotient operation, so that one is left with $F_{\underline{w}}$.
\endproof

With these definitions of the trees $T_v$ and $T/T_v$, and forests $F_{\underline{v}}$
and $T/F_{\underline{v}}$, we then have the following structure.

\smallskip

\begin{lem}\label{lemcoprod}
The operation $\Delta: \cV(\fF_{\cS\cO_0}) \to \cV(\fF_{\cS\cO_0})\otimes \cV(\fF_{\cS\cO_0})$ 
defined on trees $T\in \fT_{\cS\cO_0}$ 
\begin{equation}\label{coprodT}
\Delta(T) = \sum_{\underline{v}} F_{\underline{v}} \otimes    (T/F_{\underline{v}})  \, ,
\end{equation}
with the sum over subforests with quotients as in as in Lemma~\ref{Fquot}, 
and extended to forests by $\Delta(F)=\sqcup_a \Delta(T_a)$ for $F=\sqcup_a T_a$, 
which we write as
\begin{equation}\label{DeltaF}
\Delta(F)=\sum_{\underline{v}} F_{\underline{v}} \otimes F/F_{\underline{v}} \, ,
\end{equation}
with subforests and quotients as in Lemma~~\ref{Fquot}.
This defines a coproduct on $\cV(\fF_{\cS\cO_0})$, which
endows $\cV(\fF_{\cS\cO_0})$ with the structure of an associative,
commutative, coassociative, non-cocommutative bialgebra $(\cV(\fF_{\cS\cO_0}),\sqcup, \Delta)$.
\end{lem} 

\proof The multiplication given by disjoint union is both associative and commutative, in the case
of non-planar forests. 
We check that the coassociativity 
\begin{equation}\label{coassoc}
({\rm id}\otimes \Delta)\circ \Delta=(\Delta\otimes {\rm id})\circ \Delta
\end{equation} 
of the coproduct \eqref{coprodT} is verified. We have
$$ (\Delta\otimes {\rm id})\circ \Delta (T)=(\Delta\otimes {\rm id}) \sum_{\underline{w}} F_{\underline{w}} \otimes T/F_{\underline{w}} =
\sum_{\underline{v},\underline{w}} F_{\underline{v}} \otimes (F_{\underline{w}}/F_{\underline{v}}) \otimes T/F_{\underline{w}}\, , $$
where the sums are over subforests with the disjointness condition as in Lemma~\ref{Fquot},
where the subforests $F_{\underline{v}} \subset F_{\underline{w}}$ consists of either full components
of $F_{\underline{w}}$ or of subforests of the components. The first case gives terms of the form
$F_{\underline{v}} \otimes F_{\underline{u}} \otimes T/ F_{\underline{v},\underline{u}}$ for
$\underline{w}=(\underline{v},\underline{u})$. 
On the other hand, we have
$$ ({\rm id}\otimes \Delta)\circ \Delta (T)=({\rm id}\otimes \Delta) \sum_{\underline{v}} F_{\underline{v}} \otimes T/F_{\underline{v}}  =
\sum_{\underline{u},\underline{v}} F_{\underline{v}} \otimes (T/F_{\underline{v}})_{\underline{u}} \otimes (T/F_{\underline{v}})/(T/F_{\underline{v}})_{\underline{u}} \, . $$
We distinguish among these terms the case where the subtrees of $T$
with root at $u_i$ are disjoint from the trees of $F_{\underline{v}}$, where we
have $(T/F_{\underline{v}})/(T/F_{\underline{v}})_{\underline{u}}=T/F_{\underline{v},\underline{u}}$,
and the remaining cases where some vertices $u_i$ in $\underline{u}$, as
vertices of $T$, are above some vertices $v_j$ of the components of $F_{\underline{v}}$, in which
case the corresponding quotient is $(T/T_{v_j})/T_{u_i}=T/T_{u_i}$ and $(T/T_{v_j})_{u_i}=T_{u_i}/T_{v_j}$.
Thus, we see that we obtain the same two types of terms with the same counting. 

For a vector space $\cV$ let $\tau: \cV^{\otimes 4}\to \cV^{\otimes 4}$ denote the permutation
of the two central factors, 
\begin{equation}\label{tau}
\tau(X_1\otimes X_2\otimes X_3\otimes X_4)=X_1\otimes X_3\otimes X_2\otimes X_4\, .
\end{equation}
Multiplication and comultiplication satisfy the compatibility, 
\begin{equation}\label{prodcoprod}
 \Delta \circ \sqcup = (\sqcup \otimes \sqcup) \circ \tau \circ (\Delta\otimes\Delta), 
\end{equation} 
since $\Delta(T \sqcup T')=\Delta(T)\sqcup \Delta(T')= 
\sum_{\underline{v},\underline{v}'} F_{\underline{v}} \sqcup F'_{\underline{v}'} \otimes (T/F_{\underline{v}}) \sqcup (T'/F'_{\underline{v}'}) =(\sqcup \otimes \sqcup) \circ 
\tau  \sum_{\underline{v}} F_{\underline{v}} \otimes  (T/F_{\underline{v}}) \otimes \sum_{\underline{v}'} F'_{\underline{v}'} \otimes (T'/F'_{\underline{v}'})$.
It is convenient to include in the spanning set of $\cV(\fF_{\cS\cO_0})$ the unit $1$
of the product given by the trivial (empty) forest, which also spans the range of the counit
of the coproduct. Moreover, the vector space is graded by the
number of leaves (the length of sentences), 
$\cV(\fF_{\cS\cO_0})=\oplus_{n\geq 0} \cV(\fF_{\cS\cO_0})_n$, 
and the product and coproduct are compatible with the grading, in the sense that
$\sqcup: \cV(\fF_{\cS\cO_0})_n\otimes \cV(\fF_{\cS\cO_0})_m \to \cV(\fF_{\cS\cO_0})_{n+m}$
and $\Delta(T)=T\otimes 1 + 1 \otimes T +\sum T' \otimes T''$ where $T',T''$ are of  strictly lower 
degree than $T$, and similarly for forests. Thus, the antipode of the Hopf algebra structure 
can be defined inductively by $S(T)=-T-\sum S(T') T''$. 
\endproof

\smallskip

We can write the coproduct $\Delta$ of \eqref{coprodT} in the form
\begin{equation}\label{DeltanT}
 \Delta(T)=\sum_{n\geq 2} \Delta_{(n)}(T)\, , 
\end{equation} 
where the terms $\Delta_{(n)}$ involve extraction and quotient of
subforest of size (number of components) $n-1$, namely
with terms of the form $F_{\underline{v}}\otimes T/F_{\underline{v}}$ with
$v=(v_1,\ldots, v_{n-1})$. For any given $T$ the expression \eqref{DeltanT} is
a finite sum. In particular the first term
\begin{equation}\label{Delta2T}
\Delta_{(2)}(T) =\sum_v T_v \otimes T/T_v 
\end{equation}
corresponds to the extraction of subtrees (including the case of the trivial tree $1$ and of
the full tree $T$). Note that \eqref{Delta2T} does not suffice for coassociativity, for which
the full coproduct \eqref{coprodT} is needed, which is a form of the usual coproduct by 
admissible cuts on Hopf algebras of rooted trees. The leading term \eqref{Delta2T}
will be the relevant one for the Merge operation. 

\medskip
\subsection{Action of Merge on Workspaces}\label{ActMergeSec}

We introduce the action of Merge on workspaces by introducing an operator
that performs a search for matching terms. This will be applied to the terms
of the coproduct, that is, to the accessible terms that Merge applies to.
Indeed, the left-hand-side of the coproduct produces the list of accessible terms, 
over which the search runs, while the right-hand-side of the coproduct keeps track 
of the corresponding cancellation of copies. We will introduce the action of Merge
with some preliminary steps.

\smallskip

Suppose then given two syntactic objects, that is, two $S,S'\in \fT_{\cS\cO_0}$. 
We define a linear operator 
$$ \delta_{S,S'}: \cV(\fF_{\cS\cO_0}) \otimes \cV(\fF_{\cS\cO_0}) \to \cV(\fF_{\cS\cO_0}) \otimes \cV(\fF_{\cS\cO_0}) $$
by defining it on generators in the following way.   Let 
\begin{equation}\label{subquotF}
\fF^\Delta_{\cS\cO_0} =\{ (F_1, F_2)\in \fF_{\cS\cO_0} \times \fF_{\cS\cO_0} \, |\, 
\exists F \in \fF_{\cS\cO_0}, \, F_{\underline{v}}\subset F\, :\, F_1=F_{\underline{v}} \text{ and }
F_2=F/F_{\underline{v}}  \} \, . 
\end{equation}
For $F_1, F_2 \in \fF_{\cS\cO_0}$, we set 
\begin{equation}\label{deltaSS0}
\delta_{S,S'} (F_1\otimes F_2) = 0 \ \  \text{ for } \ \ (F_1,F_2)\notin \fF^\Delta_{\cS\cO_0}\, . 
\end{equation}
For $(F_1=F_{\underline{v}},F_2=F/F_{\underline{v}})\in \fF^\Delta_{\cS\cO_0}$ with
$F=\sqcup_{i\in \cI} T_i$, we set
\begin{equation}\label{deltaSS}
\delta_{S,S'} (F_{\underline{v}}\otimes F/F_{\underline{v}}) 
= S \sqcup S'  \otimes T_a/S \sqcup T_b/S' \sqcup F^{(a,b)} 
\end{equation}
with $F^{(a,b)}=\sqcup_{i\neq a,b} T_i$, if there are indices $a,b \in \cI$ 
such that $T_{a,v_a}\simeq S$, $T_{b,v_b}\simeq S'$. If there is more
than one choice of indices $a,b$ for which matching pairs $T_{a,v_a}\simeq S$, $T_{b,v_b}\simeq S'$
exist, then the right-hand-side of \eqref{deltaSS} should be replaced by the sum over all the
possibilities. We do not write that out explicitly for simplicity of notation.
In all other cases (where
no matching terms for $S$ and $S'$ are found) we set
\begin{equation}\label{deltaSS1}
\delta_{S,S'} (F_{\underline{v}}\otimes F/F_{\underline{v}})  = 1\otimes F \, .
\end{equation}

\smallskip

Next observe that the operation \eqref{SOMeq} on syntactic objects factors through
the grafting operator $B^+$ on forests.

\begin{defn}\label{Bplusop}
Let $\fT^\N_{\cS\cO_0}$ denote the set of all $n$-ary finite rooted trees with arbitrary $n\in \N$, 
with no assigned planar structure and with labels labelled by the set $\cS\cO_0$. Let
$\fF^\N_{\cS\cO_0}$ be the set of finite forests with connected components in $\fT^\N_{\cS\cO_0}$.
Let $\cV(\fT^\N_{\cS\cO_0})$ and $\cV(\fF^\N_{\cS\cO_0})$ denote the $\Q$-vector spaces
spanned by these sets. The grafting operator $B^+: \cV(\fF^\N_{\cS\cO_0})\to \cV(\fT^\N_{\cS\cO_0})$
is the linear operator defined on generators by
\begin{equation}\label{Bplus}
B^+(T_1\sqcup T_2 \sqcup \cdots \sqcup T_N) =  \Tree [ $T_1$   $T_2$  $\cdots$  $T_N$ ] \, .
\end{equation}
\end{defn}

The grafting operator $B^+$ is well known in the mathematical formulation of
perturbative quantum field theory, as it is the operator that defines the recursive
structure of Dyson--Schwinger equations, see \cite{BergKr}, \cite{Foissy}. 

\begin{lem}\label{MandBplus}
The Merge operator $\fM$ of \eqref{SOMeq}, namely the multiplication operation in the magma
${\rm Magma}_{na,c}(\cS\cO_0,\fM)$, determines a bilinear operator
$\fM : \cV(\fT_{\cS\cO_0}) \otimes \cV(\fT_{\cS\cO_0}) \to \cV(\fT_{\cS\cO_0})$
defined on generators as
\begin{equation}\label{MTTprime}
 \fM: T\otimes T' \mapsto \fM(T,T')= \Tree[ $T$  $T'$ ] \, ,
\end{equation} 
where we set $\fM(1,1)=1$ and $\fM(T,1)=\fM(1,T)=T$.
This operator factors through the grafting operator $B^+$ restricted to
the range of multiplication $\sqcup$, namely the diagram
$$ \xymatrix{  \cV(\fT_{\cS\cO_0}) \otimes \cV(\fT_{\cS\cO_0})  \ar[rr]^{\fM} \ar[dr]_{\sqcup} & & \cV(\fT_{\cS\cO_0}) \\ & \cV(\fF_{\cS\cO_0}) \ar[ur]_{B^+} & } $$
\end{lem}

\proof Since trees and forests do not have an assigned planar structure, both $\fM$ and the
operator $B^+$ do not depend on the order of the trees. Moreover, since the image
of $cV(\fT_{\cS\cO_0}) \otimes \cV(\fT_{\cS\cO_0})$ under $\sqcup$ consists of
forests with two connected components, so that their image under $B^+$ is still
a binary tree, which is of the form \eqref{MTTprime}.
\endproof

\smallskip

The operation \eqref{deltaSS} in combination with the operation \eqref{SOMeq} on syntactic objects 
and the bialgebra structure on workspaces contribute to the definition
of the action of Merge on workspaces as described in \cite{Chomsky17}, \cite{Chomsky19}, which
we can define in the following way. 
 
 \smallskip
 
 \begin{defn}\label{MergeWSdef}
 The action of Merge on workspaces consists of a collection of operators 
 $$\{ \fM_{S,S'} \}_{S,S'\in \fT_{\cS\cO_0}}\, ,  \ \ \  \fM_{S,S'} :  \cV(\fF_{\cS\cO_0})\to  \cV(\fF_{\cS\cO_0})\, , $$ parameterized by pairs $S,S'$ of
 syntactic objects, which act on $\cV(\fF_{\cS\cO_0})$ by
 \begin{equation}\label{MergeWSeq}
 \fM_{S,S'} = \sqcup \circ (B^+  \otimes {\rm id}) \circ \delta_{S,S'} 
 \circ  \Delta  \, ,
 \end{equation}
 with $B^+$ the grafting operator of Definition~\ref{Bplusop}. 
  \end{defn}
  
 \smallskip
 
 Note that by the definition of $\delta_{S,S'}$ the operator $B^+\otimes {\rm id}$
 applied to elements of the form $\delta_{S,S'}(F_{\underline{v}}\otimes F/F_{\underline{v}})$, 
 for $F\in \fF_{\cS\cO_0}$, produces 
 elements $X\otimes Y$ with $X$ in $\fT_{\cS\cO_0}$ and $Y$ in $\fF_{\cS\cO_0}$, hence $\fM_{S,S'}$
 maps $\fF_{\cS\cO_0}$ to itself. 
 
 \smallskip
 
 The expression \eqref{MergeWSeq} agrees with the description of the
 action of Merge on workspaces in \cite{Chomsky17}, \cite{Chomsky19},
 namely the Merge operator $\fM_{S,S'}$ searches for copies of the
 syntactic terms $S$ and $S'$ in the accessible terms of a given workspace $F$,
 extracts those accessible terms to perform the Merge operation on, and
 cancels copies from the workspace, producing the new resulting workspace. 
 
 In \eqref{MergeWSeq} the first operation, the coproduct $\Delta$, 
 produces the list of all the accessible terms $T_v$ that can be used by Merge and of the 
 corresponding remaining terms $T/T_v$ where the cancellation of copies of
 accessible terms is performed. Note that these terms correspond to just the
 part $\Delta_{(2)}$ of the coproduct as in \eqref{Delta2T}, since it is in this
 part that the nontrivial terms selected by the operator $\delta_{S,S'}$ reside,
 which searches for matching terms among the accessible terms. If no
 matching terms are found, the action is the identity and the resulting workspace
 is not changed. If matching terms are found, they are merged using $\fM$
 (or equivalently $B^+$). The final application
 of the product $\sqcup$ produces the new resulting workspace.
 The trees in the initial workspace $F$ that do not contain a pair of accessible 
 objects matching the pair $(S,S')$ remain unchanged in the workspace, 
 in agreement with the formulation of \cite{Chomsky17}, \cite{Chomsky19},
 while the trees that contain matching accessible terms are replaced by a new
 syntactic object given by merging the matching terms and by cancellation
 of the deeper copies, in the form 
  \begin{equation}\label{newWS}
  \sum_{v,w\,: T_v=S, T_w=S'} \fM(T_v,T_w) \sqcup (T/T_v) \sqcup (T/T_w) \, .
 \end{equation} 
 We describe more in detail in \S \ref{FormsMerge}
 the various cases.

 \subsection{Forms of Merge and Minimal Search}\label{FormsMerge}
 
 One of the drawbacks of the formulation \eqref{MergeWSeq} of Definition~\ref{MergeWSdef} is
 that it allows for additional forms of Merge, besides internal and external Merge, which are not
 desirable in linguistic terms, such as ``sideward Merge" or ``countercyclic Merge". We discuss
 here how a simple modification of Definition~\ref{MergeWSdef} that incorporates a formulation of
 ``Minimal Search" suffices to eliminate these cases and retain only the linguistically desirable
 cases of External and Internal Merge. This is the usual argument given in linguistics, where
 only External and Internal Merge are retained based on Minimal Search, except that here
 we reformulate it in a way that fits our algebraic setting. First we review how the different cases of Merge
 are incorporated in \eqref{MergeWSeq}, then we describe how Minimal Search is implementable
 in our Hopf algebra setting, and then we show that this has the effect of only retaining the correct
 forms of external and internal Merge.
 
 \subsubsection{Different forms of Merge}\label{ExtraMergeSec}
 
 In the description of Merge in Definition~\ref{MergeWSdef} we can distinguish several
 cases. We recall here the various cases, and we show how they occur in the
 formulation given above.
 
 Two syntactic objects $\alpha,\beta\in \cS\cO=\fT_{\cS\cO_0}$ can occur in a
workspaces $F\in \fF_{\cS\cO_0}$ either as elements (that is, as connected components
of the forest $F$), or as accessible terms of some elements. We write $T \in F$ to
indicate that a certain syntactic object $T \in \cS\cO$, seen as a tree, is a connected
component of the forest $F$.  We write $T \in Acc(T')$ to indicate that $T$ occurs
as an accessible term $T'_v$ of a syntactic object $T'\in F$.

Thus, we have the following three possibility 
\begin{enumerate}
\item $\alpha=T_i$ and $\beta=T_j$ with $T_i,T_j\in F$ and $i\neq j$;
\item $\alpha=T_i\in F$ and $\beta\in Acc(T_j)$ for some $T_j\in F$, with two
sub-cases:
\begin{itemize}
\item[a)] $i=j$
\item[b)] $i\neq j$
\end{itemize}
\item $\alpha\in Acc(T_i)$ and $\beta\in Acc(T_j)$ for some $T_i,T_j\in F$, 
with two
sub-cases:
\begin{itemize}
\item[a)] $i=j$
\item[b)] $i\neq j$
\end{itemize}
\end{enumerate}

 Case (1) describes External Merge: for a workspace 
$F=\sqcup_a T_a$, the Merge operation $\fM_{T_i, T_j}$ replaces 
the pair $T_i, T_j$ of elements of $F$ with a new syntactic object given by the 
tree $T_{ij}=\{ T_i, T_j \}=\fM(T_i,T_j)$, and produces the new workspace 
$$ F'= T_{ij} \sqcup \bigsqcup_{a\neq i,j} T_a   \, , $$
where the two components $T_i, T_j$ of $F$ have been removed and replaced by
the new tree $T_{ij}=\{ T_i, T_j \}$. External Merge decreases by one the number of
syntactic objects, and increases by two the number of accessible terms, by adding
$T_i$ and $T_j$ to the set $Acc(T_i) \cup Acc(T_j)$. In the case of 
External Merge the following is immediately evident.

\begin{lem}\label{extMergecase}
External Merge is achieved by the operators $\fM_{T_i,T_j}$ of Definition~\ref{MergeWSdef}
when the syntactic objects (trees) $T_i,T_j$ match two different connected components
of the workspace $F=\sqcup_a T_a$.
\end{lem}

Case (2a) describes Internal Merge: in this case the new workplace $F'$ 
contains a new component of the form $\fM( \beta, T_i / \beta )$ for
$\beta\in Acc(T_i)$ an accessible term of $T_i$ and $T_i$ a component of the given workspace
$F$. The quotient $T_i / \beta$ to indicate that the deeper copy of $\beta$ 
as an accessible term of $T_i$ is no longer an accessible term of $\fM( \beta, T_i / \beta )$ 
in $F'$ as $\beta$ already occurs as accessible term in a higher level in the
new syntactic object  $\fM( \beta, T_i / \beta )$ formed by Merge.
In this case, the realization of
Internal Merge by the operators $\fM_{S,S'}$ of Definition~\ref{MergeWSdef}
is more interesting, as it involves a composition of two such operators and
the role of the multiplicative unit $1$ of the Merge magma, given by the trivial tree.

\begin{prop}\label{intMergecase}
Internal Merge is realized by the operators of Definition~\ref{MergeWSdef}
as a composition $$\fM_{T/\beta,\beta} \circ \fM_{\beta,1}, $$ where $1$ is the unit of the Merge magma,
where the tree $\beta$ is an accessible term of a connected component of $F$ isomorphic to $T$.
\end{prop}

\proof The operator $\fM_{\beta,1}$ acting on the workspace $F$ will act on a term
of the form $\beta\otimes T/\beta$ in $\Delta(F)$, producing two new components 
$\beta=\fM(1,\beta)$ and $T/\beta$ in the resulting new workspace. 
The operator $\fM_{T/\beta,\beta}$ can then be applied to
these two components, as an external Merge, and the resulting workspace will now
contain the resulting term $\fM(\beta, T/\beta)$ which is the internal Merge.
\endproof

\begin{rem}\label{intextsame} {\rm Note that in this formulation internal
Merge is just repeated application of two external Merge operations, one
of them involving the magma unit, which has the effect of extracting an
accessible term and adding it, together with the cancellation of its deeper
copy, to the new workspace. However, we will see that by Minimal Search,
as well as by counting of size and number of accessible terms,
the operation $\fM_{\beta,1}$ in fact can only occur in the combination 
$\fM_{T/\beta,\beta} \circ \fM_{\beta,1}$ that is Internal Merge and not on its own. }
\end{rem}

Case (2b) corresponds to a case of Sideward Merge. In this case one obtains 
in the new workspace $F'$ a component of the form $\fM( T_i, \beta )$ and a
component of the form $T_j/ \beta$. 
Similarly, case (3b) also represents a case of Sideward Merge where in the resulting workspace $F'$
one has new components 
$\fM( \alpha, \beta )$, as well as $T_i / \alpha$ and $T_j / \beta$.
These cases of Sideward Merge also occur in the formulation of Merge of Definition~\ref{MergeWSdef},
as the following statement clearly shows.

\begin{lem}\label{SidewardMerge}
The two cases of Sideward Merge (2b) and (3b) are realized by the Merge operators of
\eqref{MergeWSeq} with $\fM_{T_i,\beta}$ with $T_i$ occurring as a component of $F$ and
$\beta$ as an accessible term of a different component $T_j$ of $F$, and $\fM_{\alpha,\beta}$
with $\alpha\in Acc(T_i)$ and $\beta\in Acc(T_j)$, for two components $i\neq j$ of $F$.
\end{lem}

The last remaining case (3a) corresponds to what is called Countercyclic Merge. In this
case the new workspace $F'$ contains new components $\fM( \alpha, \beta)$ and
$T_i / (\alpha,\beta)$, 
where we write $T_i / (\alpha,\beta)$ for the cancellation from the accessible terms 
of the copies of $\alpha$ and $\beta$ inside $T_i$. This type of Merge can also
be obtained through our formulation \eqref{MergeWSeq}. In this case, as for the
Internal Merge, one uses a composition of operators $\fM_{S,S'}$.

\begin{lem}\label{Counercyclic}
Countercyclic Merge is realized by a composition $\fM_{\alpha,\beta}\circ \fM_{\alpha,1} \circ \fM_{\beta,1}$
for $\alpha$ and $\beta$ accessible terms of the same component $T_i$ of $F$, with $\alpha$ seen
as an accessible term of $T_i/\beta$ after the application of the first $\fM_{\beta,1}$ operation.
\end{lem}

\proof This is analogous to the Internal Merge case, with the first operator $\fM_{\beta,1}$ acting
on a term $\beta \otimes T_i/\beta$ in the coproduct $\Delta(F)$,
giving rise to two new components $\beta$ and $T_i/\beta$ in the new workspace. The second
operator $\fM_{\alpha,1}$ can then be applied to this new workspace. This acts on the term of
the form $\alpha \otimes (T_i/\beta)/\alpha$, producing a new workspace that contains 
components of the form
$\alpha$, $\beta$, and $T_i / (\alpha,\beta)=(T_i/\beta)/\alpha$. 
The last operation $\fM_{\alpha,\beta}$ then acts as
External Merge on the components $\alpha$, $\beta$ of this workspace, producing
the desired new component $\fM( \alpha, \beta)$, as well as retaining the component 
$T_i / (\alpha,\beta)$ with the cancellation of the deeper copies. 
\endproof

This brief discussion shows that, in addition to Internal and External Merge, our construction
allows for extensions of Merge, such as Sideward and Countercyclic Merge that are not
desirable from the linguistic perspective. We show in the next subsection how one can
introduce a simple modification of the Merge operation described in \eqref{MergeWSeq}
that will retain only Internal and External Merge. This will make use of the grading of
the Hopf algebra, to introduce a Minimal Search that eliminates the other
extensions of Merge.

\smallskip

To avoid misunderstandings as to the purpose of the
construction in the coming \S \ref{minsearchSec} and
\S \ref{IntExtMergeSec}, it is worth stressing that our
goal here is simply to implement the usual mechanism
by which, in linguistics, only Internal and External Merge
are retained, and not other extensions of Merge, namely
the mechanism of {\em Minimal Search}. Where our presentation
differs from the usual formulation in linguistics, is that we
provide a somewhat different-looking, but in fact equivalent,
description of Minimal Search. The reason why we introduce
a reformulation of Minimal Search is the following. We are
arguing here that all the key properties of Merge follow
directly from underlying Hopf-algebraic properties.
In particular, in order to show that this is the case, we
need to reformulate all the necessary aspects of the
linguistic description of the key Merge operation in
such algebraic terms, including Minimal Search. 
This requires describing the ``minimality" property of 
Minimal Search in terms of a minimization procedure that
can be made sense of entirely in terms of the 
algebraic structure. We argue in \S \ref{minsearchSec} and
\S \ref{IntExtMergeSec} below that this can indeed be done,
with minimality expressed in the form of extraction of
leading order, with respect to a suitable grading (cost function)
associated to the terms of the coproduct. 
 
 \subsubsection{Minimal search}\label{minsearchSec}

In the formulation of Merge in \cite{Chomsky17}, \cite{Chomsky19}, the search
for matching copies of $S,S'$ in the workspace components and accessible
terms, for the application of $\fM_{S,S'}$ is performed according to a ``Minimal Search"
principle, according to which accessible terms in the higher levels of trees are
preferentially searched, before those occurring in the deeper levels. 

\smallskip

In the formulation given in
 \eqref{MergeWSeq} this Minimal Search principle is implicitly built in, through the
 structure of the coproduct. 
 Our coproduct extracts the entire list of accessible terms (simultanously implementing 
 the cancellation of copies). However, we can introduce a weight that keeps track of the
 depth of the accessible terms in each term of the coproduct. This can be done by 
 introducing formal parameters $\epsilon$ and $\eta$ and assigning to the subtrees 
 $T_v\subset T$ a weight $\epsilon^{d_v}$, where $d_v$ is the distance of the vertex $v$ 
 from the root of $T$, and to the corresponding quotient trees $T/T_v$ a weight $\eta^{d_v}$.
 This can be done by modifying the coproduct to
 \begin{equation}\label{Deltaepsiloneta}
  \Delta^{(\epsilon,\eta)} : \cV(\fT_{\cS\cO_0}) \to \cV(\fT_{\cS\cO_0})[\epsilon]\otimes_\Q 
 \cV(\fT_{\cS\cO_0})[\eta]\, , 
 \end{equation}
 $$ \Delta^{(\epsilon,\eta)}(T) =\sum_{\underline{v}} \epsilon^{d_{\underline{v}}} \, F_{\underline{v}} \otimes \eta^{d_{\underline{v}}} (T/F_{\underline{v}})\, ,  $$
 where for $\underline{v}=(v_1,\ldots, v_n)$ a set of vertices $v_i\in V_{int}(T)$, we
 set $d_{\underline{v}}=d_{v_1}+\cdots+d_{v_n}$, with $d_v$ the distance to 
 the root of $T$ as above. 
 
 \smallskip
 
 With this simple bookkeeping device, we see that, for instance, for small $\epsilon$ the higher levels 
 of the tree $T$ (internal vertices closer to the root),
 carry the largest weight, while the deeper levels are lower order contributions that
 can be neglected when one looks at the dominant terms in the coproduct. 
 Note that the introduction of the parameters $\epsilon$, $\eta$ does not alter the 
 coassociativity property of the coproduct and the compatibility with multiplication
 of Lemma~\ref{lemcoprod}.  We also assign weight $\epsilon^d$ and $\eta^d$ to the primitive
 terms of the coproduct $1\otimes T$ and $T \otimes 1$, with $d=0$ for both of these
 terms. Thus, the limit for $\epsilon \to 0$ and $\eta\to 0$ of the weighted coproduct 
 $\Delta^{(\epsilon,\eta)}(T)$ retains only the primitive part $\Delta^{(0,0)}(T)=1\otimes T + T\otimes 1$.

 \subsubsection{Internal and External Merge}\label{IntExtMergeSec}
 
 We now introduce a simple modification of the Merge operation described in \eqref{MergeWSeq},
 based on the form of Minimal Search described above, that will retain only External
 and Internal Merge, eliminating the other forms of Sideward and Countercyclic Merge.
 
 \begin{prop}\label{MSintextMerge}
Consider the modification of \eqref{MergeWSeq} given by
\begin{equation}\label{epsMergeWSeq}
 \fM^\epsilon_{S,S'} = \sqcup \circ (\fM^\epsilon  \otimes {\rm id}) \circ \delta_{S,S'} 
 \circ \Delta^{(\epsilon,\epsilon^{-1})}  \, ,
 \end{equation}
 with $\Delta^{(\epsilon,\epsilon^{-1})}$ as in \eqref{Deltaepsiloneta}, and with
 \begin{equation}\label{Mergepsilon}
 \begin{array}{c}
 \fM^\epsilon : \cV(\fT_{\cS\cO_0})[\epsilon,\epsilon^{-1}]\otimes_\Q 
 \cV(\fT_{\cS\cO_0})[\epsilon, \epsilon^{-1}] \to \cV(\fT_{\cS\cO_0})[\epsilon,\epsilon^{-1}] \\[3mm]
 \fM^\epsilon(\epsilon^d \alpha, \epsilon^\ell \beta)=\epsilon^{|d+\ell|}\, \fM(\alpha,\beta)\, . 
  \end{array}
 \end{equation}
 Then taking compositions of operations of the form \eqref{epsMergeWSeq} followed by
 evaluation at $\epsilon\to 0$ retains only External and Internal Merge
 and eliminates all other extended forms of Merge, such as Sideward and Countercyclic.
 \end{prop}
 
 \proof For a single application of \eqref{epsMergeWSeq}, one obtains terms of
 the form $\epsilon^{d_v+d_w} \fM(T_v,T_w)$, hence 
 the only terms
 remaining after taking $\epsilon\to 0$ are of the form $\fM(T,T')$ with $T,T'$
 two connected components of the workspace $F$, which have degree zero
 in the $\epsilon$ variable.  These are the External Merge cases. For a
 composition of two operators of the form \eqref{epsMergeWSeq}, we
 regard the result of the first $\fM^\epsilon_{S,S'}$ applied to a forest $F\in \fF_{\cS\cO_0}$
 as a new workspace, which now carries a dependence on the parameter $\epsilon$.
 We write such workspaces as $F(\epsilon)=\sqcup_a \epsilon^{d_a} T_a$ in the direct sum
 (as $\Q$-vector spaces) $\oplus_a \cV(\fT_{\cS\cO_0})[\epsilon, \epsilon^{-1}]$. 
 The composition with a second operator of the form \eqref{epsMergeWSeq}, then 
 produces terms of the form $\fM(\epsilon^{d_a} \epsilon^{d_{v_a}} T_{a,v_a}, 
 \epsilon^{d_b} \epsilon^{d_{v_b}} T_{b,w_b})$ for
 components $T_a, T_b$ and for vertices $v_a,w_b$ in these components.
 Thus the remaining terms of this composition, after setting $\epsilon\to 0$ will
 be those with $d_a+d_b+d_{v_a}+d_{w_b}=0$. Since $d_{v_a}+d_{w_b}\geq 0$,
 We need $d_a+d_b\leq 0$ and exactly matching the quantity $-(d_{v_a}+d_{w_b})$.
 This requires that at least one of the components $T_a$ and $T_b$ (say $T_a$) of the image
 of the first operation is a quotient $T_a=T/T_v$ of a component $T$ of the initial workspace,
 with $d_a=-d_v$. This implies that the first $\fM^\epsilon$ had the corresponding $T_v$ 
 as one of the two arguments. The other component $T_b$ is either an unchanged component
 $T_b=T'$ of the original workspace $F$, or again a quotient $T_b=T'/T'_w$ of a component, 
 if $T'_w$ was the other argument of the first $\fM^\epsilon$, or else it can be $T_b=1$, the
 trivial tree, so that in all cases either $d_b=0$ or $d_b=-d_w$. 
 Thus, we obtain either a Merge of the form $\fM(\fM(T',T_v), T/T_v)$ or of the
 form $\fM(T', \fM(T_v,T/T_v)$. The second case is clearly a composition of internal
 and external Merge, so it is of the desired form. The first case does not appear to
 be consisting only of internal/external Merge, but in fact it can be equivalently
 realized as $\fM(\fM(T',T_v),\tilde T/\fM(T',T_v))$, for another tree $\tilde T$, which
 is obtained by contracting $T_v$ to its root vertex in $T$ and gluing that root vertex
 to the root of $\fM(T',T_v)$. Thus, it is also a composition of internal and external Merge.
 The case of repeated compositions can be analyzed in the same way.
 \endproof
 
 \smallskip
 
 Thus, Proposition~\ref{MSintextMerge} shows that Minimal Search is
 equivalently described by taking the leading order term for $\epsilon \to 0$
 of the operations formed using $\fM^\epsilon_{S,S'}$.

\subsection{Other linguistic properties}

We verify here that the action of Merge on workspaces defined as in \eqref{MergeWSeq}
satisfies the desired linguistic properties. First that it accounts for the usual types of Merge:
external and internal Merge, and a form of sideway Merge. We then show that several
properties that are imposed empirically on Merge are in fact naturally built into the mathematical
formulation. In particular, we discuss the requirement that Merge does not decrease
 the total size of the workspaces and increases it at most by one. We show that this is
 indeed the case for the dominant (Minimal Search) part $ \fM^\epsilon_{S,S'} |_{\epsilon =0}$
 obtained as in Proposition~\ref{MSintextMerge}, which recovers internal/external Merge,
 while violations occur when one also includes Sideward/Countercyclic Merge, 
 confirming what is known from linguistics. 
Moreover, we show that the cancellation of copies
of accessible terms in the resulting workspaces is dictated not only by
`economy principles' but also by algebraic constraints, namely by the 
coassociativity property of the coproduct.

\subsubsection{Cases of Merge and size counting}\label{SizeSec}

We now analyze the effect of the action of Merge on
workspaces in terms of the effect on the size of workspace
and on the number of accessible terms.

\smallskip

We show that this property holds for the dominant term 
(the $\epsilon=0$ term) of the Merge action of Proposition~\ref{MSintextMerge},
which gives Internal/External Merge, while it generally fails for the other
forms of Sideward/Countercyclic Merge. This recovers an observation
already known from linguistics. 

\smallskip

We first discuss the cases of External/Internal Merge, obtained as 
$\epsilon\to 0$ dominant terms as in Proposition~\ref{MSintextMerge}. 
We then discuss the other forms of Merge that are eliminated by
Minimal Search, that is, that do not occur in the $\epsilon\to 0$ limit
of Proposition~\ref{MSintextMerge}. 

\begin{prop}\label{sizeWSlem}
Under External and Internal Merge, the effect on the counting
functions of Definition~\ref{acctermsDef} is given by the following table,
where we display the difference between the counting function after
application of Merge and before. 
\begin{center}
\begin{tabular}{| l ||r|r|r|c|}
\hline 
& $b_0$ & $\# Acc$ & $\sigma$ & $\hat\sigma$ \\
\hline 
External & $-1$ & $+2$ & $+1$ & $0$ \\
\hline
Internal & $0$ & $0$ & $0$ & $0$ \\
\hline
\end{tabular} 
\end{center}
\end{prop}

\proof In the case of External Merge, for $F=\sqcup_a T_a$, we have
$F'=\fM_{S,S'}(F)$ given by
$$ F' =\fM(T_i,T_j) \sqcup \hat F^{(i,j)}\, , $$
with $T_i\simeq S$, $T_j\simeq S'$ (where $S,S'$ are assumed
to be non-trivial syntactic objects, that is, not equal to $1$), and with
$\hat F^{(i,j)}= F\smallsetminus (T_i\sqcup T_j)$ the remaining components.
Thus, the number of connected components (of syntactic objects in the
workspace) decreases by one, $b_0(F')=b_0(F)-1$. The number of
accessible terms, on the other hand, satisfies
$$ \# Acc(F')=\# V_{int}(F')= \# Acc(F) +2 \, , $$
as the two root vertices of $T_i$ and $T_j$ become internal vertices of $\fM(T_i,T_j)$,
while all the other internal vertices remain unchanged. 
The size of the workspace satisfies
$$ \sigma(F')=b_0(F')+\# Acc(F')=\# V(F') =\sigma(F) + 1 \, , $$
since we have
$$ \sigma(\fM(T_i,T_j))=\sigma(T_i)+\sigma(T_j)+1\, . $$
The size function $\hat \sigma$ on the other hand gives
$$ \hat\sigma(F')=b_0(F')+\# V(F')=b_0(F)-1 + \# V(F) +1 = \hat \sigma(F)\, , $$
hence it is a conserved quantity under External Merge.
The special case $\fM_{S,1}$, where $S'=1$ is the trivial object (empty tree)
is discussed in Remark~\ref{noemptyMerge} below.

In the case of Internal Merge, the number of connected components
(number of syntactic objects in the workspace) remains unchanged,
as Internal Merge operates on a single tree, so $b_0(F')=b_0(F)$.
When counting the number of internal (that is, non-root) vertices,
which is the number of accessible terms, 
we see that the old root of the tree $T$, which is also the root
of $T/T_v$ becomes a new non-root vertex, while in the process
of taking the quotient $T/T_v$ according to Definition~\ref{Tquot},
two vertices are identified, hence the overall change in the
number of internal vertices is zero, and so is the change in the
total number of vertices (the size $\sigma$) where we have
$$ \sigma(T_v)+ \sigma(T/T_v) +1 =\sigma(T)\, \text{ since } \, \# Acc(T_v)+ \# Acc(T/T_v) +2 = \# Acc(T)\, , $$
because $T/T_v$ is obtained by removal of $T_v$, contraction of the
edge above the root of $T_v$ and of the other edge adjacent to it at
the vertex above the root of $T_v$, so that all the vertices of $T_v$ as
well as one additional vertex of $T$ are removed to form $T/T_v$.
This gives
$$ \sigma(F')=\sigma(\fM(T_v,T/T_v))+\sigma(\hat F)=
\sigma(T_v)+\sigma(T/T_v) + 1 +\sigma(\hat F)\, . $$
Similarly, $\hat\sigma(F')=b_0(F')+\sigma(F')=b_0(F)+\sigma(F)$,
so that the size $\hat\sigma$ also remains constant. Thus, with our choice of counting
measures as in Definition~\ref{acctermsDef}, all these
quantities are preserved unchanged under Internal Merge.
\endproof

\smallskip

\begin{rem}\label{Hcount}{\rm
The fact that under External Merge the number of syntactic objects decreases by one
and the number of accessible terms increases by two, while under Internal Merge both
the number of syntactic objects and the  number of accessible terms remain the same
is consistent with the counting in \cite{FBG}. Note that if one takes the quotient $T/T_v$
in the more sense of contracting $T_v$ to its root vertex, then the number of accessible
terms in Internal Merge would increase by one: this is the counting 
considered by Riny Huijbregts. With this choice of quotient and counting, both Internal
and External Merge would increase the number of accessible terms by exactly one. 
This makes the choice appealing, but we prefer to maintain the counting as in
\cite{FBG} because taking the quotient $T/T_v$ as in Definition~\ref{Tquot} has
advantages over the simple contraction of $T_v$ to the root, in particular not
needing projections to assign a syntactic feature label to this root vertex, which
would become a leaf in the contraction quotient.}
\end{rem}

\smallskip

\begin{rem}\label{noemptyMerge}{\rm 
In the special case of a Merge $\fM_{S,1}$, where $S' =1$ is the trivial
syntactic object (empty tree), if $S$ is matched by a component tree $T_i$
of the workspace forest $F=\sqcup_a T_a$, then $\fM(T_i,1)=T_i$ so
$F'=\fM_{S,1}(F)=F$ and the operation is just the identity. If $S$ is matched by
a subtree $T_{i,v_i}$ of a component $T_i$ of $F$, then 
$$ \fM_{S,1}(F) =\fM(T_{i,v_i},1) \sqcup T_i/T_{i,v_i} \sqcup \hat F = T_{i,v_i}  \sqcup T_i/T_{i,v_i}
 \sqcup \hat F \, , $$
for $\hat F=\sqcup_{a\neq i} T_a$. In this case the number of connected components
is growing by one, as the component $T_i$ is separated into two components
$T_{i,v_i}  \sqcup T_i/T_{i,v_i}$, while the total number of vertices is decreasing
by one, since two vertices are identified in taking the quotient $T_i/T_{i,v_i}$ while
all other vertices remain unchanged. The number of accessible terms is decreasing
by two, as the root vertex of $T_v$ is now the root of a component. Thus, we have the
table 
\begin{center}
\begin{tabular}{| l ||r|r|r|c|}
\hline 
& $b_0$ & $\# Acc$ & $\sigma$ & $\hat\sigma$ \\
\hline 
$\fM_{S,1}$ & $+1$ & $-2$ & $-1$ & $0$ \\
\hline
\end{tabular} 
\end{center}
If one imposes either that 
the size of the workspace (total number of vertices) should not decrease 
or that the number of accessible terms should not decrease, then this
implies that a Merge of the form $\fM_{S,1}$ only occurs in compositions such as
$\fM_{\beta,T/\beta}\circ\fM_{\beta,1}$ that give an Internal Merge as in Proposition~\ref{intMergecase},
but not alone, since otherwise we would violate such conditions. This is
consistent with the fact that the weight in $\epsilon$ in Proposition~\ref{MSintextMerge}
also excludes the occurrence of a Merge $\fM_{S,1}$ by itself rather than in a
composition that forms an Internal Merge.}
\end{rem}

\smallskip

\begin{rem}\label{sizeb0Acc}{\rm
When we consider separately the number of connected components
(number of syntactic objects) $b_0(F)$ of the workspace, rather than
the size $\sigma(F)=b_0(F)+\# Acc(F)$ we see that, as expected,
this decreases by one under External Merge while remaining
unchanged under Internal Merge, so that the number of
components decreases overall during the course of a derivation, as expected,
leading to the desired ``convergence".
} \end{rem}

\smallskip

The remaining cases of the Merge operation \eqref{MergeWSeq},
which are subdominant in $\epsilon \to 0$ in \eqref{epsMergeWSeq}
(hence eliminated by Minimal Search) have a different behavior
with respect to the counting functions of Definition~\ref{acctermsDef}.

\begin{prop}\label{sizeWSlem2}
In the cases of Sideward and Countercyclic Merge in \eqref{MergeWSeq} 
we have the following change in the counting functions of Definition~\ref{acctermsDef}.
\begin{center}
\begin{tabular}{| l ||c|c|c|c|}
\hline 
& $b_0$ & $\# Acc$ & $\sigma$ & $\hat\sigma$ \\
\hline 
Sideward (3b)  &  $+1$ & $0$  & $+1$  & $+2$   \\
\hline
Sideward (2b) &  $0$ & $+1$  & $+1$   & $+1$   \\
\hline
Countercyclic (3a) (i) &  $+1$ & $\# Acc(T_{a,w_a})$   & $\sigma( T_{a,w_a} )$ & $ \sigma( T_{a,w_a} ) +1$  \\
\hline
Countercyclic (3a) (ii) & $+1$  & $\# Acc(T_{a,v_a})$   & $\sigma( T_{a,v_a})$  &  $\sigma( T_{a,v_a})  + 1$  \\
\hline
Countercyclic (3a) (iii) & $+1$  & $-2$  & $-1$  & $0$   \\
\hline
\end{tabular} 
\end{center}
where in case (3a) $T_{a,v_a}$ and $T_{a,w_a}$ are the two subtree of
the same component $T_a$ used for Countercyclic Merge. 
\end{prop}

\proof  In the case of case of Sideward Merge, case (3b) of Section~\ref{ExtraMergeSec},
we have, for $F=\sqcup_i T_i$, 
$$ F' = \fM(T_{a,v_a}, T_{b,w_b}) \sqcup T_a/T_{a,v_a} \sqcup T_b/T_{b,w_b}\sqcup
\hat F^{(a,b)} \, , $$
with $\hat F^{(a,b)}=\sqcup_{i\neq a,b} T_i$. Thus, the number
of connected components increases by one, because of the new
component $\fM(T_{a,v_a}, T_{b,w_b})$, while the number of accessible terms
is given by
$$ \# Acc(F')=\# Acc(\fM(T_{a,v_a}, T_{b,w_b}) )+\# Acc(T_a/T_{a,v_a}) +\# Acc(T_b/T_{b,w_b})
+\# Acc(\hat F^{(a,b)}) $$
$$ = \# Acc(T_{a,v_a}) +\,  \# Acc(T_{b,w_b}) +2 +\, \# Acc(T_a) -\, \# Acc(T_{a,v_a}) - 1 $$ $$
+\, \# Acc(T_b) -\, \# Acc(T_{b,w_b}) -1 + \, \# Acc(\hat F^{(a,b)}) = \# Acc(F)\, . $$
Thus, $\sigma(F')=b_0(F')+\# Acc(F')=\sigma(F)+1$ and $\hat\sigma(F')=\hat\sigma(F)+2$.

For Sideward Merge, case (2b) of Section~\ref{ExtraMergeSec} we similarly have
$$ F' = \fM(T_a, T_{b,w_b}) \sqcup T_b/T_{b,w_b} \sqcup \hat F^{(a,b)}\, , $$
so that $b_0(F')=b_0(F)$, since one new component $\fM(T_a, T_{b,w_b})$
is created and one component $T_b$ is removed. The counting of
accessible terms give
$$ \# Acc(F')=\# Acc(\fM(T_a, T_{b,w_b}) )+ \# Acc(T_b/T_{b,w_b} ) + \# Acc (\hat F^{(a,b)}) $$
$$ = \# Acc(T_a) + \# Acc(T_{b,w_b}) +2 + \# Acc(T_b) -  \# Acc(T_{b,w_b}) -1 + \# Acc (\hat F^{(a,b)}) 
= \# Acc(F) +1 \, . $$
Thus, the size satisfies $\sigma(F')=\sigma(F)+1$ and $\hat\sigma(F')=\hat\sigma(F)+1$.

In the case of Countercyclic Merge we have, for $F=\sqcup_i T_i$
$$ F' = \fM(T_{a,v_a},T_{a,w_a}) \sqcup T_a /T_{a,v_a,w_a} \sqcup \hat F^{(a)} \, , $$
where $T_{a,v_a,w_a}\subset T_a$ is given by 
\begin{equation}\label{Tavw}
 T_{a,v_a,w_a}:=\left\{ \begin{array}{ll}  T_{a,v_a} & \text{case (i): if } T_{a,w_a}\subset T_{a,v_a} \\[3mm]
T_{a,w_a} & \text{case(ii): if } T_{a,v_a}\subset T_{a,w_a} \\[3mm]
T_{a,v_a}\sqcup T_{a,w_a} & \text{case (iii): if } T_{a,v_a}\cap T_{a,w_a}=\emptyset \, ,
\end{array} \right. 
\end{equation}
and $\hat F^{(a)}=\sqcup_{i\neq a} T_i$.
Thus, we obtain one additional connected
component $\fM(T_{a,v_a},T_{a,w_a})$, so that
$$ b_0(F')=b_0(F)+1 \, . $$
The counting of accessible terms is given by
$$ \# Acc(F')= \# Acc(\fM(T_{a,v_a},T_{a,w_a})) + \# Acc(T_a /T_{a,v_a,w_a} ) + \# Acc (\hat F^{(a)}) $$
$$ = \# Acc(T_{a,v_a}) + \# Acc(T_{a,w_a}) + 2 + \# Acc(T_a /T_{a,v_a,w_a} ) + \# Acc (\hat F^{(a)})\, , $$
because the root vertices of $T_{a,v_a}$ and $T_{a,w_a}$ also appear as accessible terms in
$\fM(T_{a,v_a},T_{a,w_a})$, 
while we have
$$ \# Acc (F) = \# Acc(T_a) + \# Acc (\hat F^{(a)}) = \# Acc(T_{a,v_a}) + 2 + \# Acc(T_a/T_{a,v_a}) 
+ \# Acc (\hat F^{(a)}) $$
$$ = \# Acc(T_{a,w_a}) + 2 + \# Acc(T_a/T_{a,w_a}) 
+ \# Acc (\hat F^{(a)}) \, . $$
Thus, in the first two cases (i) and (ii) of \eqref{Tavw}, we respectively have
$$ \# Acc(F')=\left\{ \begin{array}{ll}
\# Acc(F) +  \# Acc(T_{a,w_a})       & \text{if } T_{a,v_a,w_a} = T_{a,v_a}  \\[3mm]
\# Acc(F) +  \# Acc(T_{a,v_a})        & \text{if } T_{a,v_a,w_a} = T_{a,w_a}    
\end{array}\right. $$
This then gives, for these two cases
$$ \sigma(F')= \left\{ \begin{array}{ll}
\sigma(F)+ \sigma(T_{a,w_a}) & \text{if } T_{a,v_a,w_a} = T_{a,v_a}  \\[3mm]
\sigma(F)+ \sigma(T_{a,v_a}) & \text{if } T_{a,v_a,w_a} = T_{a,w_a}    \, ,
\end{array}\right. $$
since $\sigma(F')=b_0(F')+\# Acc (F')$, 
while $\hat\sigma(F')=b_0(F')+\sigma(F')$ has an additional increase by $+1$. 

The third case (iii) of \eqref{Tavw} has two possibilities. If there is a vertex $u_a$ in $T_a$ 
that is adjacent to both the roots of $T_{a,v_a}$ and $T_{a,w_a}$ then the
tree $T_a$ contains a subtree with root $u_a$ that is isomorphic to $\fM(T_{a,v_a},T_{a,w_a})$
so that we have, in this case,
$$ F'= \fM(T_{a,v_a},T_{a,w_a})\sqcup T_a/\fM(T_{a,v_a},T_{a,w_a}) \sqcup \hat F^{(a)} \, , $$
so that the counting
of accessible terms satisfies
$$ \# Acc (F') = \# Acc(\fM(T_{a,v_a},T_{a,w_a})) + \# Acc(T_a/\fM(T_{a,v_a},T_{a,w_a})) +
\# Acc(\hat F^{(a)})$$ $$ = \# Acc(T_a) - 2 + \# Acc(\hat F^{(a)})  = \# Acc(F)-2 \, . $$
The workspace sizes correspondingly change by
$$ \sigma(F')=b_0(F')+ \# Acc (F') =\sigma(F) -1 \ \ \text{ and } \ \ 
\tilde\sigma(F')= 2 b_0(F')+ \# Acc (F') = \tilde\sigma(F) \, . $$
The other possibility for case (iii) is that the vertices above the roots of $T_{a,v_a}$ and $T_{a,w_a}$
are different. In this case, $T_a/T_{a,v_a,w_a}=(T_a/T_{a,v_a})/T_{a,w_a}=(T_a/T_{a,w_a})/T_{a,v_a}$.
Thus, we have
$$ \# Acc(T_a/T_{a,v_a,w_a}) + \# Acc(T_{a,w_a}) + 2 = \# Acc(T_a/T_{a,v_a}) $$
$$ = \# Acc (T_a) - \# Acc(T_{a,v_a}) -2 \, , $$
so that
$$ \# Acc(T_a/T_{a,v_a,w_a}) = \# Acc (T_a) -  \# Acc(T_{a,v_a}) -  \# Acc(T_{a,w_a}) -4 \, . $$
Thus we have
$$ \# Acc(F')= \# Acc(\fM(T_{a,v_a},T_{a,w_a}))+ \# Acc(T_a/T_{a,v_a,w_a}) + \# Acc(\hat F^{(a)}) $$
$$ = \# Acc(T_{a,v_a}) +  \# Acc(T_{a,w_a}) + 2 +  \# Acc(T_a/T_{a,v_a,w_a}) + \# Acc(\hat F^{(a)}) $$
$$ = \# Acc(T_a) -2 + \# Acc(\hat F^{(a)}) = \# Acc(F) -2 \, , $$
so that we obtain the same counting as in the first case of (iii). 
\endproof

\smallskip

We see from Proposition~\ref{sizeWSlem2} that various requirements on the
counting functions of Definition~\ref{acctermsDef} can be used to rule out
these forms of Sideward and Countercyclic Merge. In particular we look
at the effect of requirements that the number of accessible terms in the workspace should be
non-decreasing ($\Delta (\# Acc) \geq 0$)  and the number of syntactic objects should be 
non-increasing ($\Delta b_0 \leq 0$);
that the overall size of the workspace does not decrease and does
not increase more than one ($0\leq \Delta \sigma \leq 1$), 
and the requirement that $\hat\sigma$ is a  conserved quantity ($\Delta \hat\sigma=0$).
Note that all of these conditions are satisfied by Internal and External Merge.

\begin{cor}\label{remainCases}
Constraints on counting functions in the cases of 
Sideward and Countercyclic Merge are (Y) or are not (N)
satisfied according to the following table.

\medskip
\begin{center}
\begin{tabular}{| l || c | c | c | c |}
\hline 
  & $\Delta b_0 \leq 0$ & $\Delta (\# Acc) \geq 0$ & $0\leq \Delta \sigma \leq 1$ & $\Delta \hat\sigma=0$ \\
\hline
Sideward (3b)  & N   &  Y & Y  & N \\
\hline
Sideward (2b) &  Y  & Y & Y & N  \\
\hline
Countercyclic (3a) (i) & N & Y & N & N   \\
\hline
Countercyclic (3a) (ii) &N  & Y & N & N  \\
\hline
Countercyclic (3a) (iii) & N & N & N & Y   \\
\hline
\end{tabular}
\end{center}
\medskip

Thus, all the cases of Sideward and Countercyclic Merge of 
Proposition~\ref{sizeWSlem2} are ruled out by at least one
of these conditions, but not all of them by the same one,
and $\Delta b_0 \leq 0$ and $\Delta (\# Acc) \geq 0$ together
do not suffice to rule out all of these cases, but
conservation $\Delta \hat\sigma=0$ together with any one
of the other conditions suffices.
\end{cor}

This shows that constraints based on counting are less efficient than the
constraint based on Minimal Search (the $\epsilon \to 0$ limit in Proposition~\ref{MSintextMerge}),
which rules out all of these remaining forms of Sideward and Countercyclic Merge.

\begin{rem}\label{Accplusone}{\rm 
Note however that, if one take quotients by contraction to the root vertex, instead
of using the quotient as in Definition~\ref{Tquot}, then the condition that
{\em the number of accessible terms should increase exactly by one} (as suggested
by Huijbregts, see Remark~\ref{Hcount}) would suffice to rule out all the cases of 
Sideward and Countercyclic Merge. Indeed, 
with the quotient by contraction, we would obtain that the number 
of accessible terms for both Sideward and Countercyclic Merge would always
increase by at least two. }\end{rem}

\smallskip

\subsubsection{Cancellation of copies}\label{CopiesSec}

In the form \eqref{MergeWSeq} of the action of Merge on workspaces, the
cancellation of copies of the accessible terms used by Merge is implemented
by the coproduct $\Delta$ of \eqref{coprodT} through
the quotient terms $T/T_v$. 

\smallskip

Cancellation of copies is usually postulated as an ``economy principle" 
in linguistics, and it is usually assumed that cancellation always happens
in the deeper copies. 

\smallskip

A first observation is that, in the formalism we are using, the fact
that cancellation is implemented in the deeper copy is directly built
into the structure of the coproduct and it does not have to be 
included as an additional requirement, since in the terms $T_v \otimes T/T_v$
the copy of $T_v$ on the left-hand-side (the one that contributed to Merge)
has lower depth than the copy inside $T$, which is cancelled on the right-hand-side.

\smallskip

A second observation is that cancellation of copies is necessary in order
to have a good coassociative coproduct. Indeed, one needs to quotient out 
the copy of $T_v$ inside $T$ in the right-hand-side of the coproduct for
coassociativity to work as shown in Lemma~\ref{lemcoprod}. One can
see that a coproduct of the form $T\mapsto \sum_v T_v \otimes T$ without
the cancellation would no longer have this property.

\smallskip

Note moreover that, although we refer to the term $T/T_v$ as ``cancellation"
of a copy of $T_v$, nothing is really cancelled, since a copy of $T_v$
remains on the other side of the term $T_v \otimes T/T_v$ of the coproduct.
The basic structure of the coproduct separates out trees (in all
possible ways) into a subtree and a quotient.
One can simply then read the subtree as the ``creation of a copy"
and the quotient tree as corresponding ``cancellation of
the original (deeper) copy" when Internal Merge is applied.

\smallskip

Also observe that there are distinct roles in the model we are discussing
here for copies and repetitions. Repetitions are accounted for in this setting
by the fact that we are defining the workspace as a forest (a disjoint union
of trees, that is, of syntactic objects). This allows for the presence of
repetitions since a forest is not a set but a multiset of trees. Copies,
on the other hand are only created during the application of the
coproduct, and ultimately play a role only in the operation of Internal Merge.

\smallskip

\section{The core computational structure of Merge}\label{CoreSec}

The description of Merge and its action on workspaces that we described above 
follows closely the formulation presented in \cite{Chomsky17}. We discuss here a
further simplification of the structure of Merge, which extract its core computational
structure, as presented in \cite{Chomsky21b}. 

\smallskip

 Let $\fT$ be the set of binary rooted trees without planar structure (and without
 labeling of the leaves), and $\cV(\fT)$ the free $\Z$-module (or the $\Q$-vector
 space) spanned by the set $\fT$. The following description is the analog of
 Definition~\ref{SOdef} and Remark~\ref{remSOtrees}.

 \begin{lem}\label{nassAlg}
 The set $\fT$ is the free non-associative, commutative magma whose elements are
 the balanced bracketed expressions in a single variable $x$, with the binary
 Merge operation $(\alpha,\beta)\mapsto \fM(\alpha,\beta)=\{ \alpha, \beta \}$. Correspondingly,
 $\cV(\fT)$ is the free commutative non-associative algebra generated by a
 single variable $x$.
 \end{lem}
 
 \proof We can identify the binary rooted trees without planar structure with
 the balanced bracketed expressions in a single variable $x$. For example
 $$ \{ \{ x \{ x x \} \} x \} \longleftrightarrow \Tree [ [ x  [ x x ] ]  x ] \ \ \, . $$
 The Merge operation $\fM(\alpha,\beta)=\{ \alpha, \beta \}$ takes two
 such bracketed expressions $\alpha$ and $\beta$ and forms a new one
 of the form $\{ \alpha, \beta \}$, which correspond to attaching the roots of the
 two binary trees to a common root, $\fM(T,T')=T\wedge T'$. 
 \endproof

 Equivalently, $\cV(\fT)$ is the free algebra over the quadratic operad
 freely generated by the single
commutative binary operation $\fM$ (see \cite{Holt3}).

\smallskip

The generative process for the set $\fT$ via the Merge operation can
be equivalently described as a recursive procedure encoded in the form
of a fixed point equation.

\begin{prop}\label{propDStreesMerge}
Let $\cV(\fT)=\oplus_\ell \cV(\fT)_\ell$ with the grading by length (number of leaves) as before,
with $\fM: \cV(\fT)_\ell \times \cV(\fT)_{\ell'} \to \cV(\fT)_{\ell + \ell'}$, where $\fM$ is
extended by linearity in each variable. Consider formal infinite sums
$X=\sum_{\ell \geq 1} X_\ell$ with $X_\ell \in \cV(\fT)_\ell$ and the recursive equation
\begin{equation}\label{DStreesMerge}
X = \fM(X,X) \, .
\end{equation}
Then the generative process for $\fT$ via the Merge operation is equivalent
to the recursive construction of a solution of \eqref{DStreesMerge} with initial 
condition $X_1=x$. 
\end{prop}

\proof
We have
$\fM(\sum_\ell X_\ell, \sum_{\ell'} X_{\ell'}):=\sum_{\ell,\ell'} \fM(X_\ell, X_{\ell'})$.
In particular, the term of degree $n$ in $\fM(X,X)$ is given by
$$ \fM(X,X)_n = \sum_{j=1}^{n-1} \fM(X_j,X_{n-j}) \, , $$
so that the fixed point equation \eqref{DStreesMerge} reduces to the recursive 
relation
$$ X_n = \sum_{j=1}^{n-1} \fM(X_j,X_{n-j})\, . $$
starting with $X_1=x$, the recursion produces $X_2=\{ x x \}$, 
$X_3 = \{ x \{ x x \}\} + \{ \{ x x \} x \} = 2 \{ x \{ x x \}\}$, $X_4=2 \{ x \{ x \{ x x \}\} \} +
\{ \{ x x \} \{ x x \} \}$, and so on. These first terms $X_n$ list all the possible non-planar
binary rooted trees with $n$ leaves, with multiplicities that account for the 
different planar structures. Given a non-planar binary rooted tree $T$ with $n$ leaves,
we can always write it as $T=\fM(T',T'')$ where $T',T''$ are the two binary rooted
trees with roots at the two internal vertices of $T$ connected to the root of $T$, with $j=\# L(T')$
and $n-j=\# L(T'')$, for some $j\in \{ 1, \ldots, n-1 \}$. Since
the Merge product is commutative it does not matter in which order we list $T'$ and $T''$. 
Thus, each $T\in \fT_n$ can be mapped uniquely to an {\em unordered} pair $\{ T', T'' \}$
and conversely, any pair of trees $T', T''$ with numbers of leaves $\ell'$ and $\ell''$, respectively,
determines uniquely a tree $\fM(T',T'')$ with $\ell'+\ell''$ leaves. Thus, 
inductively, if each $X_j$ for $1\leq j<n$ consists of a list (formal sum) of all the
possible non-planar binary rooted tress with $j$ leaves, then $X_n$ also consists of
a sum of all the possible non-planar binary rooted tress with $n$ leaves. One sees
similarly that the integer coefficients in the sum count different planar structures.
\endproof

\smallskip

As we discuss further in \S \ref{PhysSec}, this shows that the generative process
for the core computational structure of Merge in the Minimalist Model of
syntax is in fact the most fundamental basic case of the Dyson--Schwinger
equations in physics. We give a quick summary here of what we will discuss
in more detail in \S \ref{PhysSec}. 

\smallskip

In general, the Dyson--Schwinger equation implements in perturbative quantum field
theory the construction of solutions of the equations of motion. It is a way of encoding
the variational principle of least action for equations of motion
in classical physics in a form suitable for quantum fields, via a recursive method of
solution that can be performed order by order in the perturbative expansion. There are two main
conceptual aspects to single out here. One is the fact that the construction
of solutions of Dyson--Schwinger equations becomes a combinatorial problem,
in terms of Feynman graphs and associated trees, expressible as a solution to a 
fixed point equation, of which \eqref{DStreesMerge} is the most fundamental example. 
The general such combinatorial Dyson--Schwinger equation always involves
a form of (possible $n$-ary) Merge operation, given by the grafting operator $B^+$ of
Definition~\ref{Bplusop}, and a polynomial fixed point equation
in a Hopf algebra, which takes the general form
$X =B^+(P(X))$, for a polynomial $P$ and a variable $X=\sum_\ell X_\ell$ in (a completion of)
a Hopf algebra of rooted trees. The equation is solved recursively, as in the fundamental 
case of \eqref{DStreesMerge}. The other aspect is the usual requirement that for classical 
solutions of the equations of motion the action functional is stationary under infinitesimal variations.
This is transformed in the case of quantum fields into corresponding equations for the quantum
correlation functions. In the formulation of
perturbative quantum field theory in terms of Hopf algebra, these in turn arise 
from the combinatorial solution, which is entirely determined in terms of the
underlying Hopf-algebraic structure, together with the evaluation of a (renormalized)
Feynman rule, to obtain the actual physical solution from the combinatorial one.
As we discuss further in \S \ref{PhysSec}, the first observation identifies the
generative process of syntactic objects through Merge with the basic case of the structure
of generative processes of fundamental physics. The second observation suggests that
the optimality that the core computational structure of Merge ought to satisfy is of the
same conceptual nature as the least action principle of physics, when the latter manifests
itself in a combinatorial form.

\smallskip

\section{Constraints on Merge: the $n$-arity question}

An important question regarding the Merge operation of syntax is whether
the same generative power would be achievable with a similar operation
that is $n$-ary, for some $n\geq 3$, rather than binary. 

\smallskip

Riny Huijbregts presented in \cite{Huij} 
strong empirical linguistic evidence  for why, for example, a ternary Merge 
would be inadequate, in the sense that such a ternary operation would 
produce both {\em undergeneration} and {\em overgeneration} with
respect to the binary Merge. Undergeneration refers to syntactic
constructions that can be derived through the binary Merge but would
not be generated by a ternary Merge, while overgeneration consists
of ungrammatical sentences that would be generated by a ternary Merge,
but not by binary Merge. While the undergeneration problem could in
principle by bypassed by hypothesizing the simultaneous presence of
a binary and a ternary Merge, the overgeneration problem cannot be
similarly dealt with. 

\smallskip

We discuss here briefly why any
$n$-ary Merge operation, for any $n\geq 3$, would necessarily lead
to both undergeneration and overgeneration, as a simple consequence
of the algebraic structure described in the previous section. In particular,
within this formulation one can see that undergeneration and
overgeneration have two somewhat different origins. Undergeneration
is a direct consequence of the structure of the magma on the
Merge operation, which gives rise to the set of syntactic objects, while
overgeneration involves directly the action of Merge on workspaces.

\medskip
\subsection{The $n$-ary Merge magma}

Here we assume the existence of a hypothetical $n$-ary Merge,
for some $n\geq 3$, and we discuss how the structure of the
magma of syntactic objects changes with respect to the binary case.
We assume the same initial set $\cS\cO_0$ of lexical terms and
syntactic features.

\smallskip

\begin{defn}\label{nmagma}
An $n$-magma consists of a set $X$ together with an $n$-ary operation
$$ \fM_n: X \underbrace{\times \cdots \times}_{n\text{-times}} X \to X\, , \ \ \ \  ( x_1, \ldots, x_n ) \mapsto \fM_n(x_1,\ldots, x_n) \, . $$
We say that $(X, \fM_n)$ is an $n$-magma over a set $Y$, if all elements
of $X$ are obtained by iterated application of $\fM_n$ starting with $n$-tuples of elements in $Y$.
\end{defn}

\smallskip

We write $\{ x_1,\ldots, x_n \}:=\fM_n(x_1,\ldots, x_n)$ for the element of $X$ that is
obtained by applying $\fM_n$ to the $n$-tuple $( x_1, \ldots, x_n )$. In particular,
the set $X$ consists of a subset $X_1$ consisting of all elements of the form
 $\{ y_1,\ldots, y_n \}:=\fM_n(y_1,\ldots, y_n)$ with all the $y_i\in Y$, a set $X_{2n-1}$
 consisting of all elements of the form
 $$ \fM_n( y_1,\ldots, y_{i-1},  \fM_n(a_{i,1},\ldots, a_{i,n})  , y_{i+1}, \ldots, y_n )= 
 \{ y_1,\ldots, y_{i-1},  \{ a_{i,1},\ldots, a_{i,n} \}  , y_{i+1}, \ldots, y_n \}$$
 for $i=1,\ldots, n$ and with all the $y_i, a_{i,j} \in Y$, a set $X_{3n-2}$ consisting of
 all elements of the form
 $$ \{ y_1,\ldots, y_{i-1},  \{ a_{i,1},\ldots, a_{i,n} \}  , y_{i+1}, \ldots, y_{j-1},  
 \{ b_{j,1},\ldots, b_{j,n} \}   , y_{j+1}, \ldots y_n \}\, \, \text{ and } $$
 $$ \{ y_1,\ldots, y_{i-1},  \{ a_{i,1},\ldots, a_{i,j-1},    \{ b_{j,1},\ldots, b_{j,n} \}    , a_{i, j+1}, a_{i,n} \}  , y_{i+1}, \ldots,  y_n \}\, ,  $$
 with $i\neq j$, $i,j=1,\ldots, n$, and all the $y_i, a_{i,k}, b_{j,k} \in Y$, and so on, so that we have
 \begin{equation}\label{gradenmagma}
  X= \bigsqcup_{k\geq 1} X_{k(n -1)+1} \, . 
 \end{equation}
  We refer to the subset $X_{k(n -1)+1}$ as the set of elements of length $k(n -1)+1$ in the $n$-magma. 
  
 \smallskip

The $n$-magma is associative if all the elements of length $k(n -1)+1$ are
identified, that is, if bracketing is irrelevant. It is commutative if elements $\{ x_1,\ldots, x_n \}$ with entries
that differ by a permutation in the symmetric group $S_n$ are identified, that is, if every set within
brackets is unordered. 

\smallskip

We then have the following description of the set of syntactic objects produced by
a hypothetical $n$-ary Merge $\fM_n$.

\smallskip

 \begin{defn}\label{defSOn}
 The set $\cS\cO^{(n)}$ of $n$-ary syntactic objects is the free, non-associative, commutative $n$-magma
 on the set $\cS\cO_0$,
 \begin{equation}\label{magmaSOn}
 \cS\cO^{(n)}={\rm Magma}^{(n)}_{na,c}(\cS\cO_0, \fM_n) \, ,
 \end{equation}
 with 
 \begin{equation}\label{SOnSet}
 \cS\cO^{(n)}= \bigsqcup_{k\geq 1} \cS\cO^{(n)}_{k(n -1)+1} \, . 
\end{equation} 
 \end{defn}
 
 \smallskip
 
 \begin{rem}{\rm
 We can identify the elements of $\cS\cO^{(n)}$ with rooted $n$-ary trees,
 \begin{equation}\label{nSOtrees}
\cS\cO^{(n)}  \simeq \fT^{(n)}_{\cS\cO_0}\, ,
\end{equation}
namely trees where all the non-leaf vertices have $n$ descendants, without
a planar structure, and with leaves labelled by elements of the set $\cS\cO_0$.
(Note that what we call here $n$-ary trees are  {\em full}
$n$-ary trees.)
 }\end{rem}
  
 \smallskip
 
 The set $\cS\cO^{(n)}_{k(n -1)+1}$ is the set of rooted $n$-ary trees
(with no assigned planarity) with $k(n -1)+1$ leaves, and therefore with $k$ non-leaf
vertices. The number $k$ of non-leaf vertices is the number of applications of
$\fM_n$ in the process of generating elements of $\cS\cO^{(n)}$, where each
non-leaf vertex is the graphical representation of a Merge operation. 

\medskip
\subsection{Undergeneration}\label{undergenSec}

Given the structure \eqref{SOnSet} of the set of $n$-ary syntactic objects,
we can show that there are two different forms of undergeneration (with
respect to the binary Merge), and that both of them inevitably occur for
any $n$-ary Merge with $n\geq 3$. The two different forms of undergeneration
correspond, respectively, to certain lengths not being achievable through an 
$n$-ary Merge construction, and to certain syntactic parsing ambiguities not
being accountable for by an $n$-ary Merge construction.

\smallskip

The first form of undergeneration can be seen as follows. 

\begin{lem}\label{undergen1}
Only strings of elements of $\cS\cO_0$ of length $k(n -1)+1$, for some $k\geq 1$,
can be achieved through an $n$-ary Merge. In particular, only the binary Merge
can achive all lengths.
\end{lem} 

\proof The number of leaves of an $n$-ary tree with $k$ non-leaf
vertices is $k(n -1)+1$. Thus, the only possible strings of elements of $\cS\cO_0$
that can be obtained through $k$ successive applications of an $n$-ary Merge
$\fM_n$ are of length $k(n -1)+1$, as in the decomposition \eqref{SOnSet} 
of the set of $n$-ary syntactic objects. Only in the case $n=2$ the set
$\{ k(n -1)+1 \}_{k \geq 1}$ contains all positive integers greater than or
equal to $2$.
\endproof

\smallskip

Known empirical linguistic examples of this kind of undergeneration 
include, for instance, the fact that sentences like ``{\em it rains}"
are in $\cS\cO_2$ while $\cS\cO^{(3)}_2=\emptyset$.

\smallskip  

The second form of undergeneration can be seen through counting
and comparing the sizes of the sets $\cS\cO_{k(n -1)+1}$ and 
$\cS\cO^{(n)}_{k(n -1)+1}$, for $n\geq 3$. 
The counting formulae for rooted trees are simpler in the case
of trees with an assigned planar structure, rather than for abstract
trees with no assigned planarity. Thus, we count the resulting 
trees after the externalization step that introduces planar structures.

\smallskip

Let $\fT_{\cS\cO_0}^{pl}=\sqcup_\ell \fT_{\cS\cO_0, \ell}^{pl}$ and $\fT_{\cS\cO_0}^{(n),pl}=\sqcup_k \fT_{\cS\cO_0, k(n -1)+1}^{pl}$ denote, respectively, the sets of binary and of $n$-ary
rooted trees with a choice of planar embedding.

\smallskip

\begin{lem}\label{undergen2}
For any given $n\geq 3$, and for $\ell=k(n -1)+1$, for any $k\geq 2$, we have
$$ \# \fT_{\cS\cO_0, \ell}^{pl} > \# \fT_{\cS\cO_0, \ell}^{pl} \, . $$
\end{lem}

\proof
The number of planar rooted binary trees with
$\ell=r+1$ leaves (hence $r$ non-leaf vertices) is given by the Catalan number
$$ C_r =\frac{1}{r+1} \binom{2r}{r} \, . $$
Thus, for $\ell=k(n -1)+1$, we have the counting
$$ C_{k(n -1)} =\frac{1}{(n -1)k+1} \binom{2k(n -1)}{k}\, . $$
The number of planar rooted $n$-ary trees with $(n-1)k+1$ leaves (hence $k$
non-leaf vertices) is correspondingly given by the Fuss--Catalan numbers
$$ C^{(n)}_k =\frac{1}{(n-1)k+1} \binom{nk}{k} \, . $$
The different assignments of labels at the leaves contribute in both cases
a factor $S^{(n-1)k+1}$, where $S:=\# \cS\cO_0$.
When we compare the counting we see that
\begin{equation}\label{Catalanrel}
 S^{(n-1)k+1} (C_{k(n -1)}-C^{(n)}_k)= \frac{S^{(n-1)k+1}}{(n -1)k+1} \left(  \binom{2k(n -1)}{k} - \binom{nk}{k}\right) > 0 
 \end{equation}
since $2k(n-1)>nk$ for $n\geq 3$. 
\endproof

\smallskip

Thus, at the level of planar trees, counting 
detects an undergeneration phenomenon which is present at all levels $k\geq 1$
of the construction of the sets of syntactic objects. 
This phenomenon shows that there are always strings of 
elements of $\cS\cO_0$ of length $k(n -1)+1$ that have
ambiguous parsing when realized in terms of binary Merge, while the
ambiguity cannot be accounted for with an $n$-ary Merge.

\smallskip

As a simple example of this type of undergeneration, the two different
parsings of the ambiguous sentence ``I saw someone with a telescope"
depend on the difference between the two binary trees 
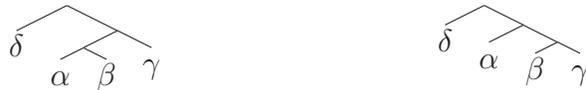
\begin{center}
\Tree [ $\delta$ [ [ $\alpha$  $\beta$ ] $\gamma$ ] ]   \ \ \    
\Tree [ $\delta$ [ $\alpha$ [ $\beta$ $\gamma$ ] ] ]
\end{center}
which would disappear entirely if the terms $\alpha,\beta,\gamma,\delta$
are assembled through a $4$-ary Merge to form the tree 
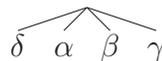
\begin{center}
\Tree [ $\delta$  $\alpha$  $\beta$  $\gamma$ ]   \ \ \ \  \ \ \ \  \ \ \ \  \ \ \ \  \ \ \ \ 
\end{center} 
where the ambiguity would no longer be detectable. 

\medskip
\subsection{The structure of a hypothetical $n$-ary Merge}

Given the set $\cS\cO^{(n)}$ of syntactic objects associated to a hypothetical $n$-ary
Merge, obtained as in \eqref{SOnSet}, we can consider the same type of action of
Merge on workspaces that we have introduced above for a binary Merge. We will see in \S \ref{OvergenSec}
below that, when the same structure is implemented through an $n$-ary Merge with $n\geq 3$,
it inevitably leads to an overgeneration phenomenon.

\smallskip

As in the binary case, we introduce the set of workspaces as finite collections of syntactic
objects, which in the $n$-ary case are elements of the set $\cS\cO^{(n)}\simeq \fT_{\cS\cO_0}^{(n)}$
of $n$-ary non-planar rooted trees. We again consider the vector space $\cV(\fF_{\cS\cO_0}^{(n)})$,
where $\fF_{\cS\cO_0}^{(n)}$ is the set of finite forests with connected components in $\fT_{\cS\cO_0}^{(n)}$.
In order to write the extraction of accessible terms and the cancellation of copies in the form
of a coproduct, and the Merge pairing on accessible terms, 
we consider the relevant algebraic structure on $\cV(\fF_{\cS\cO_0}^{(n)})$, namely 
the product given by disjoint union $\sqcup$ and the coproduct as in \eqref{coprodT}
and \eqref{DeltaF}.

\smallskip

\begin{rem}\label{Tquotnary}{\rm 
In defining a coproduct of the form \eqref{coprodT} for
an $n$-ary tree, one no longer has the option of taking the quotient $T/T_v$ in
the sense of Definition~\ref{Tquot}, as after removal of the subtree $T_v$, contractions
of edges in the resulting tree $T\smallsetminus T_v$ will produce vertices with either
less or more than $n$ descendants. In order to have a quotient $T/T_v$ that is
itself an $n$-ary tree, one can define $T/T_v$ as obtained by contracting $T_v$ to its
root vertex. This requires that the root vertex of $T_v$, which becomes a
leaf in $T/T_v$, need to be labelled by an element in $\cS\cO_0$. This requires
including in $\cS\cO_0$ 
syntactic features of the form $XP$ with $X\in \{ N, V, A, P, C, T, D, \ldots \}$,
and the label of the new leaf in  $T/T_v$, obtained by projection,  needs to
be computed by inspecting the structure of $T_v$. This computation of
labels of root vertices can be avoided in the case of binary Merge, by
performing quotients as in Definition~\ref{Tquot}, but can no longer be avoided
in the case of $n$-ary Merge with $n\geq 3$, where the quotient needs to
be taken by contraction to the root vertex. }
\end{rem}

\smallskip

We can assume that the form of the action of Merge on workspaces will be of the same
form as in the binary case of \eqref{MergeWSeq}. Thus, we can write the desired
form for the $n$-ary Merge action on workspaces as follows.

\smallskip

Given a collection $S=(S_i)_{i=1}^n$ of $n$-ary syntactic objects $S_i\in \fT^{(n)}_{\cS\cO_0}$,
we also define an operator 
$$ \delta_{S_1,\ldots, S_n} :  \cV(\fF^{(n)}_{\cS\cO_0})\otimes \cV(\fF^{(n)}_{\cS\cO_0})
\to  \cV(\fF^{(n)}_{\cS\cO_0})\otimes \cV(\fF^{(n)}_{\cS\cO_0}) $$ 
in the same way as the $\delta_{S,S'}$ defined in the binary case in \eqref{deltaSS0}, 
\eqref{deltaSS}, \eqref{deltaSS1}.  As in \eqref{subquotF} we set
\begin{equation}\label{subquotFn}
\fF^{\Delta,(n)}_{\cS\cO_0} =\{ (F_1, F_2)\in \fF^{(n)}_{\cS\cO_0} \times \fF^{(n)}_{\cS\cO_0} \, |\, 
\exists F \in \fF^{(n)}_{\cS\cO_0}, \, F_{\underline{v}}\subset F\, :\, F_1=F_{\underline{v}} \text{ and }
F_2=F/F_{\underline{v}}  \} \, .
\end{equation}
We then set
\begin{equation}\label{deltaSn0}
\delta_{S_1,\ldots, S_n} (F_1\otimes F_2)=0 \ \ \ \text{ for } (F_1, F_2)\notin \fF^{\Delta,(n)}_{\cS\cO_0} \, ,
\end{equation}
\begin{equation}\label{deltaSn}
\delta_{S_1,\ldots, S_n}( F_{\underline{v}}, F/F_{\underline{v}} ) = S_1\sqcup \cdots \sqcup S_n \otimes
T_{a_1}/S_1 \sqcup \cdots \sqcup T_{a_n}/S_n \sqcup F^{(a_1,\ldots, a_n)} \, , 
\end{equation}
for $F=\sqcup_{i\in \cI} T_i$, if there are indices $a_1,\ldots, a_n \in \cI$ such that $S_i \simeq T_{a_i, v_i}$,
and with $$F^{(a_1,\ldots, a_n)}=\sqcup_{i\neq a_1,\ldots, a_n} T_i.$$ 
As in the binary case, if there is more
than one choice of indices $a_1,\ldots, a_n$ for which matching terms $S_i \simeq T_{a_i, v_i}$
exist, then the right-hand-side of \eqref{deltaSn} should be replaced by the sum over all the
possibilities, which we do not write out explicitly.  In the remaining case where
such matching of terms does not exist one sets
\begin{equation}\label{deltaSn1}
\delta_{S_1,\ldots, S_n}( F_{\underline{v}}, F/F_{\underline{v}} ) =1 \otimes F\, .
\end{equation} 

\smallskip

\begin{defn}\label{nMergeWSdef}
The action of Merge on workspaces consists of a collection of operators 
 $$\{ \fM_{S_1,\ldots, S_n} \}_{S_i '\in \fT^{(n)}_{\cS\cO_0}}\, ,  \ \ \  \fM_{S_1,\ldots, S_n} :  \cV(\fF^{(n)}_{\cS\cO_0})\to  \cV(\fF^{(n)}_{\cS\cO_0})\, , $$ parameterized by $n$-tuples $(S_i)_{i=1}^n$ of
 $n$-ary syntactic objects, which act on $\cV(\fF^{(n)}_{\cS\cO_0})$ by
 \begin{equation}\label{nMergeWSeq}
 \fM_{S_1,\ldots, S_n} = \sqcup \circ (B^+  \otimes {\rm id}) \circ \delta_{S_1,\ldots, S_n} 
 \circ \Delta \, ,
 \end{equation}
 with the same operation $B^+$ as in Definition~\ref{Bplusop}.
\end{defn}

Note that the $n$-ary analog of Lemma~\ref{MandBplus} also holds, so that
\eqref{nMergeWSeq} is obtained analogously.

This action of Merge on workspaces has the same structure as in the binary case,
namely, for each of the $n$ input of the $n$-ary Merge $\fM_n$ a search is made
over the workspace by extracting accessible terms and comparing them with the
corresponding $n$-ary syntactic object $S_i$. Non-matching terms are left
unchanged in the new workspace, while the $n$-ary Merge operation is applied
to $n$-tuples of matching terms among the extracted accessible terms for each
Merge input. The new workspace then have these Merge outputs along with
the terms coming from the quotient  part of the coproducts, where cancellation 
of the deeper copies of the
accessible terms used by Merge is performed. 

\smallskip

One can envision other possible generalizations of the binary Merge
action on workspaces to the $n$-ary case, using a coproduct with
higher arity instead of $\Delta$.
We will not discuss them here, since \eqref{nMergeWSeq} is the simplest
direct generalization of \eqref{MergeWSeq}, and it suffices to show the
 inevitability of overgeneration (which would occur for the same reasons
 in other generalizations as well).

\medskip
\subsection{Overgeneration}\label{OvergenSec}
 
 We can now see the 
overgeneration phenomenon as a different type of comparison between
the sets $\cS\cO$ and $\cS\cO^{(n)}$, with respect to the undergeneration
discussed above. Unlike undergeneration, overgeneration depends not only
 on the structure of the set $\cS\cO^{(n)}$ of syntactic objects, but also on the action of
 on workspaces as described above. 
 
 \smallskip
 
 Indeed, consider the following empirical linguistic example of overgeneration by
a hypothetical ternary Merge. 
We take a workspace given by an $n$-ary forest of the form
$$ F=   \{ \alpha, \beta, \gamma \} \sqcup \delta \sqcup \eta \, , $$
with  $\alpha,\beta,\gamma,\delta,\eta \in \cS\cO^{(3)}\simeq \fT^{(3)}_{\cS\cO_0}$
Consider the action of ternary Merge on workspaces described by \eqref{nMergeWSeq}
with $n=3$ and with $S=(S_1,S_2,S_3)$ given by 
$S_1=\alpha$, $S_2=\beta$, and $S_3=\{ \alpha, \beta, \gamma \}$ gives the internal Merge
$$ \fM_{S_1,S_2,S_3}(F) =\{ \alpha , \beta, \{ \alpha, \beta, \gamma \}\} \sqcup \delta \sqcup \eta \, . $$
Similarly, the same action with $S_1=\delta$, $S_2=\eta$, and $S_3=\{ \alpha, \beta, \gamma \}$
gives the external Merge
$$ \fM_{S_1,S_2,S_3}(F) =\{  \delta,\eta, \{ \alpha, \beta, \gamma \} \}\, . $$
These ternary Merge operations are responsible for generating ungrammatical
sentences such as\footnote{This example was communicated to us by Riny Huijbregts. For a
more detailed discussion, see \cite{Huij}.} {\em peanuts monkeys children will throw} (as opposed to  {\em 
children will throw monkeys peanuts}), resulting from 
\begin{equation}\label{example3merge}
 \{ \text{peanuts}, \text{monkeys}, \{ \text{children}, \text{will}, 
 \{\text{throw}, \text{monkeys}, \text{peanuts} \}\}\} \, . 
\end{equation} 

\smallskip

In the example of \eqref{example3merge} one sees that $\alpha$ and $\beta$ are
accessible terms of $\{ \alpha, \beta, \gamma \}$, hence with a ternary Merge one can form 
$\{ \alpha, \beta, \{ \alpha, \beta, \gamma \}\}$. On the other hand, $\{ \alpha , \beta \}$ is 
not an accessible term of  $\{ \{ \alpha , \gamma \}, \beta \}$.

\smallskip

This example indicates that the overgeneration phenomenon it illustrates is
caused by a difference in the size of the sets of accessible terms on which
the action of Merge on workspaces is based. Indeed, in the general
case of an arbitrary hypothetical $n$-ary Merge with $n\geq 3$, the
overgeneration phenomenon is caused by the simple fact that, given a binary tree 
and an $n$-ary tree with the same set of leaves, there are fewer pairs of
accessible terms (input for binary Merge) in the binary tree than there are
$n$-tuples of accessible terms (input for the $n$-ary Merge) in the $n$-ary tree.
We can see this more explicitly as follows. 

\smallskip

\begin{lem}\label{lemOvergen}
Let $V_{int}^o(T)$ denote the set of vertices that are neither leaves nor root.
Suppose given a set of leaves $L$
with $\# L=\ell=k(n-1)+1$, for some $k\geq 1$. Let $T$ and $T'$ be, respectively,
a binary and an $n$-ary tree with $L(T)=L(T')=L$. Then for $k\geq n$ we have
\begin{equation}\label{eqOvergen}
 \# (V_{int}^o(T')^{n-1} \smallsetminus {\rm Diags})  >   \# V_{int}^o(T)\, , 
\end{equation}  
where ${\rm Diags}\subset V_{int}^o(T')^{n-1}$ is the union of all the diagonals,
where two or more of the entries in $(v_1,\ldots, v_{n-1})\in V_{int}^o(T')$ coincide.
\end{lem} 

\proof A binary tree $T$ with $\ell$ leaves has $\ell-1$ non-leaf
vertices, a total of $2\ell-1$ vertices, and $2(\ell-1)$ non-root vertices. 
An $n$-ary tree on the same set of leaves with $\ell=k(n-1)+1$ has
$k$ non-leaf vertices, a total of $kn+1$ vertices, and $kn$ non-root vertices. 
Note that, in order to have $n-1$ choices without repetitions in $V_{int}^o(T')$
we need to assume that $k\geq n$. 
In  \eqref{eqOvergen} we are then comparing
$\# V_{int}^o(T)=k(n-1)$ with $$ \# (V_{int}^o(T')^{n-1} \smallsetminus {\rm Diags}) =
k(k-1)\cdots (k-n+1) = n! \binom{k}{n}=\frac{k!}{(k-n)!}\, , $$
which is larger than $k(n-1)$. 
\endproof

\smallskip

We include the leaf-vertices in the counting of accessible terms. 
The result of Lemma~\ref{lemOvergen} is similar. 

\begin{cor}\label{leavesAcc}
If $V_{int}(T)$ is the set of all non-root vertices, then
\begin{equation}\label{eqOvergen2}
 \# (V_{int}(T')^{n-1} \smallsetminus {\rm Diags})  >   \# V_{int}(T)\, , 
\end{equation}  
\end{cor}

\proof Note that, unlike in Lemma~\ref{lemOvergen}, 
now $k\geq 1$ is arbitrary.  By the same counting as above of non-root vertices, in this case we have
$\# V_{int}(T)=2k(n-1)$ and $\# V_{int}(T')=kn$. In particular, $\# V_{int}(T)> \# V_{int}(T')$,
but when counting inputs for internal Merge we obtain
$$ \# (V_{int}(T')^{n-1} \smallsetminus {\rm Diags}) = 
kn (kn-1) \cdots (kn-n+1) = n! \binom{kn}{n} =\frac{(nk)!}{(n(k-1))!} $$
which is now larger than $\# V_{int}(T)$. 
\endproof

The left-hand-side of \eqref{eqOvergen} (respectively, \eqref{eqOvergen2}) is the size of the set of possible
inputs for an $n$-ary internal Merge that can be extracted from the $n$-ary tree $T'$,
while the right-hand-side of \eqref{eqOvergen} (respectively, \eqref{eqOvergen2}) 
is the size of the set of all possible inputs
for a binary internal Merge that can be extracted from the binary tree $T$, with the same
set of leaves. The discrepancy between these two sizes shows the inevitable
presence of overgeneration with an $n$-ary Merge and quantifies precisely the
amount of overgeneration that can occur. 

\section{A model of externalization} 

The action of Merge on workspaces described in \eqref{MergeWSeq} and Definition~\ref{MergeWSdef}
can be also interpreted as a representation of a non-associative algebra in the following way.
(All vector spaces and algebras are taken over $\Q$.)

\smallskip

First observe that the magma structure on $\cS\cO=\fT_{\cS\cO_0}$ of \eqref{SOeq} gives to the
vector space $\cV(\fT_{\cS\cO_0})$ the structure of a non-associative commutative algebra, see
\cite{Holt1}, \cite{Holt2}, where the binary Merge operation $\fM$ gives the product operation. 

\smallskip

Note that the coproduct \eqref{coprodT} does not induce a bialgebra structure on
$\cV(\fT_{\cS\cO_0})$ because it does not satisfy the compatibility:
$$ \Delta \circ \fM \neq (\fM\otimes \fM)\circ \tau \circ (\Delta\otimes \Delta)\, , $$
unlike the compatibility of $\sqcup$ and $\Delta$ on $\cV(\fF_{\cS\cO_0})$ in Lemma~\ref{lemcoprod}.
The reason is because $\Delta(\fM(T,T'))$ has only terms of the form
$\fM(T_v,T')\otimes T/T_v$, $\fM(T,T'_w)\otimes T'/T'_w$, $T_v\otimes \fM(T/T_v,T')$, 
and $T'_w\otimes \fM(T,T'/T'_w)$, while the right-hand-side applied to $T\otimes T'$ also
has all terms of the form $\fM(T_v,T'_w)\otimes \fM(T/T_v, T'/T'_w)$. However, a
modified form of the coproduct \eqref{coprodT}  does give $\cV(\fT_{\cS\cO_0})$ the
structure of a non-associative, commutative, co-commutaive, co-associative Hopf
algebra (see \cite{Holt1}, \cite{Holt2}), with 
\begin{equation}\label{modifyDelta}
 \Delta(T)= \sum_{L\subset L(T)} T|_L \otimes T|_{L^c} \, , 
\end{equation} 
where, for a subset $L\subset L(T)$ (with $L^c=L(T)\smallsetminus L$)
we write $T|_L$ to denote the binary rooted tree obtained by removing
all the leaves in $L$ and then performing the edge contractions needed to obtain a binary tree.
The difference between this coproduct and \eqref{coprodT} lies in the fact that the coproduct
of \eqref{modifyDelta} would correspond to a notion of accessible terms that 
include all possible subsets of the set of leaves, not just those of the form $L=L(T_v)$. 

\smallskip

Here we only need to consider the non-associative commutative algebra
structure $\cA_{na,c}=(\cV(\fT_{\cS\cO_0}), \fM)$, without the comultiplication, but
the above remark is included for completeness.

\smallskip
\subsection{Merge representation}

The notion of representation and module over a
non-associative algebra is much weaker than its associative counterpart.
If $\cA$ is a non-associative algebra and $\cV$ is a vector space, an $\cA$-module
structure on $\cV$ is simply given by a {\em linear} map
$$ \rho: \cA \to {\rm End}(\cV)\, . $$
This map is not an algebra homomorphism when $\cA$ is non-associative. 
We can equivalently view $\rho$ as a linear map $\rho: \cA \times \cV \to \cV$. 
We say that a vector space $\cV$ is a module over a non-associative algebra $\cA$ if it is
endowed with a representation of $\cA$ on $\cV$ in the sense described here above.

\smallskip

\begin{lem}\label{repAVlem}
The vector space $\cV(\fF_{\cS\cO_0})$ is a module over the algebra $\cA_{na,c}$
through the representation given by the maps
\begin{equation}\label{repAV}
\rho(T)(F) = \sqcup \circ (\fM^T \otimes 1) \circ \Delta \, (F) =\sqcup_a ( \fM(T,T_{a,v}) \sqcup T_a/T_{a,v} ) ,
\end{equation}
where $F=\sqcup_a T_a$ and $\fM^T(T_a):=\fM(T,T_a)$. 
\end{lem}

\smallskip

It then suffices to show that the representation \eqref{repAV} is enough to determine the
Merge operators $\fM_{S.S'}$ as described in \eqref{MergeWSeq} in Definition~\ref{MergeWSdef}. 
The form  \eqref{MergeWSeq} of the action of Merge on workspaces is designed so as
to exactly describe the procedure of searching among the accessible terms and syntactic objects of
the workspace for copies of the chosen objects $S$ and $S'$, for each of the two inputs of Merge,
and applying Merge with corresponding cancellation of copies of accessible terms. The following
observation shows that the same internal and external Merges can be also obtained through
the somewhat simplified expression \eqref{repAV} of the representation of Lemma~\ref{repAVlem}.

\smallskip

\begin{lem}\label{repMerge}
The representation  \eqref{repAV} suffices to determined the Merge operations \eqref{MergeWSeq}  
on workspaces in $\cV(\fF_{\cS\cO_0})$.
\end{lem}

\proof Consider the operator $\rho(T)$ of the representation \eqref{repAV} 
restricted to the subspace $\cV_T \subset \cV(\fF_{\cS\cO_0})$ spanned by
all the workspaces $F\in \fF_{\cS\cO_0}$ that contain $T$ either as a connected
component of $F$ or as an accessible term of one of the connected components.
Then the result $\rho(T)(F)$ of applying the operator $\rho(T)$ to elements
$F\in \cV_T$ gives rise to all the possible Merge operations $\fM_{S,S'}$
with $S=T$ and with $S'$ another component or accessible term of $F$.
\endproof

\smallskip

This rephrasing of the action of Merge on workspaces in terms of algebras and 
modules has the advantage that it suggests
a possible model for thinking about the process of externalization. This models how
the core computational structure of Merge, when implemented in the human brain, needs 
to be followed by what one calls an ``externalization procedure", that allows for interaction 
with the sensorimotor system (see \cite{BerCh}). It is in this externalization process that 
additional constraints are imposed, such as the presence of a linear ordering on sentences (in the 
form of planar embeddings of binary rooted trees), as well as constraints coming from UG 
principles. One also needs to account for the syntactic diversity across different human 
languages (syntactic parameters), see for instance \cite{Eve}.

\smallskip
\subsection{Externalization and linear ordering} \label{OrderSec}

We can look first at the step of externalization that introduces planar structures,
hence linear ordering on the leaves of the trees, that is, an ordering on the
resulting sentence. At first it may seem, intuitively, that introducing a linear
ordering is a way of imposing a constraint and should
therefore give rise to some kind of quotient map, in fact the quotient
map goes in the opposite direction, as the map that 
identifies the abstract (non-planar) tree behind all its different
planar embeddings. It can also be described as the
quotient that maps non-commuting variables (where order matters)
to corresponding commuting variables (where it does not).  
This means that one has a non-associative and non-commutative
algebra $\cA_{na,nc}$ together with a projection homomorphism
$\cA_{na,nc} \to \cA_{na,c}$ that quotients out the commutators
and identifies all different planar embeddings to the same 
abstract (non-planar) tree. The part of the externalization process 
that fixes a planar structures consists of the choice of a section of 
this projection morphism. Such a section is not an algebra
homomorphism (as that would not map a commutative to a
non-commutative algebra). Indeed, this is not surprising, as it is simply
expressing the fact that the choice of planar embeddings cannot be 
universal and is in fact language-dependent, as it involves specific word order
structures. Thus, the construction of this section of the projection 
$\cA_{na,nc} \to \cA_{na,c}$ is the first instance where one sees the
role of syntactic parameters, in this case specifically in the form of
word order parameters. 

\smallskip

We denote, as before, by $\fT_{\cS\cO_0}$ and $\fF_{\cS\cO_0}$ the sets of 
binary rooted trees (respectively, forests) with leaves labels in $\cS\cO_0$,
and we denote by $\fT^{pl}_{\cS\cO_0}$ and $\fF^{pl}_{\cS\cO_0}$ the corresponding 
sets of {\em planar} binary rooted trees (respectively, forests) with leaves labels in $\cS\cO_0$.

\begin{prop}\label{plTF}
At the level of the underlying vector spaces, the quotient map $\cA_{na,nc} \to \cA_{na,c}$
is the map $\Pi: \cV(\fT^{pl}_{\cS\cO_0})\twoheadrightarrow \cV(\fT_{\cS\cO_0})$
that assigns to a planarly embedded tree the underlying abstract tree, forgetting the
planar embedding, that is, identifying together all the different planar embeddings
of the same abstract tree. There is a corresponding quotient map on workspaces
$$ \cV(\fF^{pl}_{\cS\cO_0})\twoheadrightarrow \cV(\fF_{\cS\cO_0}) \, . $$
The representation  \eqref{repAV} extends to a representation $\rho^{pl}: \cA_{na,nc} \to {\rm End}(\cV(\fF^{pl}_{\cS\cO_0}))$ so that the following diagram commutes
\begin{equation}\label{plDiag}
\xymatrix{ \cA_{na,nc} \otimes \cV(\fF^{pl}_{\cS\cO_0}) \ar[r]^{\ \ \ \rho^{pl}} \ar[d]^{\Pi\otimes \Pi} & \cV(\fF^{pl}_{\cS\cO_0})\ar[d]^{\Pi} \\ 
 \cA_{na,c} \otimes \cV(\fF_{\cS\cO_0}) \ar[r]^{\ \ \rho} & \cV(\fF_{\cS\cO_0}) \, .}
\end{equation}
\end{prop}

\proof The algebra $\cA_{na,nc}$ is the free non-associative non-commutative algebra
generated by the set $\cS\cO_0$ with a non-associative non-commutative product,
which we denote by $\fM^{nc}$. Unlike the non-associative commutative Merge
product $\fM$ of $\cA_{na,c}$, we have in general $\fM^{nc}(\alpha,\beta)\neq 
\fM^{nc}(\beta,\alpha)$. Thus, we can identify  $\cA_{na,nc}$ with the algebra
associated to the non-associative non-commutative magma ${\rm Magma}(\cS\cO_0,\fM^{nc})$.
We can identify the elements of this magma with {\em ordered} words in the alphabet
$\cS\cO_0$ with matched parentheses. Equivalently, we can describe the magma 
${\rm Magma}(\cS\cO_0,\fM^{nc})$ through its Malcev representation, where 
a new variable $c$ is introduced to mark the opening parenthesis. The position of the closing
parenthesis is determined, so for example, instead of $(\alpha, ((\beta, \gamma),\delta)))$
one can write $c \alpha c^2 \beta\gamma\delta$, see \cite{Holt2}. The magma operation 
$\fM^{nc}$ in the Malcev representation takes the form
$$ \fM^{nc}(\alpha,\beta) = c \, \alpha\, \beta\, . $$
The set of ordered words in $\cS\cO_0$ with matched parentheses can be identified
with the set of binary rooted trees with a choice of planar embedding. 
Thus, we can identify
$$  \fT^{pl}_{\cS\cO_0} = {\rm Magma}(\cS\cO_0,\fM^{nc}) \, . $$
In terms of the Malcev representation, the variable $c$ marks the opening parenthesis
that corresponds to an internal vertex of the planar tree in $\fT^{pl}_{\cS\cO_0}$.
The quotient map $\cA_{na,nc} \to \cA_{na,c}$ that kills the commutators has kernel
the ideal generated by the elements $\fM^{nc}(T,T')-\fM^{nc}(T',T)$. 
Elements in this ideal are by construction differences between pairs of trees that
differ in planar embeddings at one (or more) of the internal vertices, since every
application of $\fM^{nc}$ corresponds to an internal vertex of the resulting 
planar tree. Thus, the quotient map is exactly $\Pi$ that identifies different planar
embeddings of the same tree. Similarly, in $\fF^{pl}_{\cS\cO_0}$ forests are
now planarly embedded, hence the components $T_a$ form an ordered set,
which we describe by writing $F=\sqcup^{nc}_a T_a$, where $\sqcup^{nc}$
means that the order of the $T_a$ matters, namely $\sqcup^{nc}$ is the
union as planarly embedded trees, in a sequential order compatible with
an ordering of the union of their leaves.  
By defining $\rho^{nc}$ as 
$$ \rho(T)(F) = \sqcup^{nc} \circ (\fM^{T,nc} \otimes 1) \circ \Delta \, (F) =\sqcup^{nc}_a ( \fM^{nc}(T,T_{a,v}) 
\sqcup T_a/T_{a,v} )
$$
with $F=\sqcup^{nc}_a T_a$ and $\fM^{T,nc}(T_a):=\fM^{nc}(T,T_a)$, one obtains
compatibility as expressed by the commutativity of the diagram in the statement.
\endproof 

\smallskip

An assignment of a planar structure can then be seen as a section $\sigma_L$ of the projection $\Pi$,
\begin{equation}\label{sigmaL}
 \xymatrix{  \fT^{pl}_{\cS\cO_0} \ar[r]_{\Pi} &  \fT_{\cS\cO_0}\, , \ar@/_1pc/[l]_{\sigma_L}  } 
\end{equation}  
with $\Pi\circ \sigma_L={\rm id}$, where the section is dependent on a particular language $L$
and exists as a map of vector spaces, but not as a morphism of algebras. These properties
express the property that assignment of linear ordering of sentences is not directly genereated
by Merge itself, but requires an additional mechanism, and cannot be implemented in a
universal language-independent way, see the discussion in \S \ref{ParamSec}.

\smallskip
\subsection{Correspondences} \label{CorrSec}

There is another role for syntactic parameters in the model of externalization
process we propose here, where they define a quotient map that significantly
cuts down on the combinatorial explosion of Merge. In order to describe this
process more precisely, it is useful to recall the mathematical notion of
correspondence and how it generalizes the concept of function and mapping.

\smallskip

The notion of correspondence is a natural generalization of the concept of
function or map, and has already played a crucial role in  
contemporary mathematics. It is generally 
understood that correspondences provide a better notion of morphisms 
than functions. In the case of a category of geometric spaces (or the underlying
category of sets) one typically replaces the usual notion of a function
$f: X \to Y$ with correspondences that are of the form
\begin{equation}\label{Zcorr}
 \xymatrix{ & Z \ar[dl] \ar[dr] & \\ X & & Y\, . } 
\end{equation} 
The case of a function is recovered as the special case where $Z=G(f)\subset X\times Y$
is the graph of the function $G(f)=\{ (x,y)\,|\, y=f(x) \}$, with the two projection maps
to $X$ and $Y$. Correspondences, however, are more general than functions. 
Given a correspondence $Z$, one can transfer structures (e.g. vector bundles, spaces of functions, etc.)
from $X$ to $Y$, by pulling them back to $Z$ and then pushing them forward to $Y$ via the two maps
of the correspondence.

\smallskip

Thus, in this setting, given a category $\cC$ that has pullbacks, one can view 
correspondences as $1$-morphisms
in a $2$-category of {\em spans} in $\cC$, namely the $2$-category ${\rm Spans}(\cC)$ that
has:
\begin{itemize}
\item objects given by the objects of $\cC$;
\item $1$-morphisms given by $\cC$-diagrams of the form \eqref{Zcorr}, 
with the composition given by the pullback 
$$ \xymatrix{ & & Z\times_Y Z' \ar[dl] \ar[dr] & & \\
& Z \ar[dl] \ar[dr] & & Z'  \ar[dl] \ar[dr] & \\
 X & & Y & & X' \, ; }   $$
\item $2$-morphisms between spans $X \leftarrow Z_1 \rightarrow Y$ and $X\leftarrow Z_2 \rightarrow Y$
are morphisms $Z_1 \to Z_2$ in $\cC$ that give a commutative diagram
$$  \xymatrix{ & Z_1 \ar[dd] \ar[dl] \ar[dr] & \\ X & & Y \\
& Z_2 \ar[ul] \ar[ur] &  }  $$
\end{itemize}

\smallskip

Since correspondences are usually described in this way as {\em spans} in the case of geometric
spaces, they are usually described dually as {\em cospans} in the case of algebras, namely
as diagrams of the form
$$ \xymatrix{ & \cE & \\ \cA \ar[ur] & & \cB \ar[ul]\, . } $$
This construction further extends to the typical case where correspondences of algebras are defined
as bimodules. However, one can also consider the case of
correspondences (or co-correspondences) given by spans of algebras
$$  \xymatrix{ & \cE \ar[dl] \ar[dr] & \\ \cA & & \cB \, . } $$
(and co-spans of spaces), with the composition $(\cB\stackrel{g}{\leftarrow} \cE' \rightarrow \cA')
\circ (\cA\leftarrow \cE \stackrel{f}{\rightarrow} \cB)$ given by the pullback, that is, 
the restricted direct sum $\cE\oplus_\cB \cE'=\{ (e,e')\,|\, f(e)=g(e')\}$.
This is the kind of correspondences that we see in the description of the externalization of Merge. 

\smallskip
\subsection{Externalization as correspondence} \label{ExtCorrSec}

The computational mechanism described by the action of Merge on workspaces encodes
the fundamental computational structure of syntax, which is independent of the variation of
syntactic structures across different languages. Where this variation actually occurs is only 
in the externalization process. 
At the level of the syntactic objects, given by the trees 
in $\fT^{pl}_{\cS\cO_0}$, and of the workspaces, given by the forests in $\fF^{pl}_{\cS\cO_0}$, 
the externalization that corresponds to a particular language $L$ introduce quotient maps
\begin{equation}\label{quotL}
\begin{array}{l}
\Pi_L: \fT^{pl}_{\cS\cO_0} \twoheadrightarrow \fT^{pl,L}_{\cS\cO_0} \\[3mm]
\Pi_L: \fF^{pl}_{\cS\cO_0} \twoheadrightarrow \fF^{pl,L}_{\cS\cO_0} \, ,
\end{array}
\end{equation}
where $\fT^{pl,L}_{\cS\cO_0}$ and $\fF^{pl,L}_{\cS\cO_0}$ are the set of planar binary rooted trees
(respectively, forests) with leaves labels in $\cS\cO_0$, that are possible syntactic trees for the
given language $L$. This quotient map very significantly reduces the combinatorial explosion
of Merge, as only a small fraction of all the possible binary rooted trees generated by the
Merge magma are realizable as syntactic trees of a specific given language. 
(We discuss in \S \ref{ParamSec} below the role of syntactic parameters
in determining the quotient map \eqref{quotL}.)

\smallskip

In order to formulate this quotient at the level of the algebra $\cA_{na,nc}$ and its action
$\rho^{pl}$ on the space $\cV(\fF^{pl}_{\cS\cO_0})$ of workspaces with planar structure,
we need to use the notion of {\em partial algebra}, which is a vector space induced
with a partially defined bilinear multiplication. Examples of partial algebras include the
span of paths in a directed graph with the composition product. 

\smallskip

\begin{lem}\label{subalgL}
The projection map of vector spaces $\Pi_L:  \cV(\fT^{pl}_{\cS\cO_0}) \twoheadrightarrow \cV(\fT^{pl,L}_{\cS\cO_0})$  induced by the quotient map of \eqref{quotL} determines a non-associative
non-commutative partial algebra $\cA_{na,nc,L}$, with an induced action $\rho^{pl,L}$ of 
$\cA_{na,nc,L}$ on $\cV(\fF^{pl,L}_{\cS\cO_0})$, with a commutative diagram
\begin{equation}\label{partialalgDiag}
 \xymatrix{  \cA_{na,nc,L} \otimes \cV(\fF^{pl,L}_{\cS\cO_0}) \ar[r]^{ \ \ \ \rho^{pl,L}} & \cV(\fF^{pl,L}_{\cS\cO_0}) \\
 \cA_{na,nc} \otimes \cV(\fF^{pl}_{\cS\cO_0}) \ar[r]^{\ \ \ \rho^{pl}}\ar[u]^{\Pi_L} & \cV(\fF^{pl}_{\cS\cO_0})\, .\ar[u]^{\Pi_L} 
} 
\end{equation}
\end{lem}

\proof As a vector space, $\cV(\fT^{pl,L}_{\cS\cO_0})$ is spanned by those 
trees in $\fT^{pl}_{\cS\cO_0}$ that are realizable as syntactic trees of the given language $L$,
as such it can be viewed either as a quotient space, under the projection $\Pi$ determined
by \eqref{quotL}, or as a subspace of $\cV(\fT^{pl}_{\cS\cO_0})$.  
This subspace $\cV(\fT^{pl,L}_{\cS\cO_0})$ is not a priori a subalgebra with
respect to the Merge product $\fM^{nc}$. However, it is a partial algebra, where the induced
Merge $\fM^{nc,L}$ acts as $\fM^{nc}$ on the domain given by the set of 
pairs $T,T'\in \cV(\fT^{pl,L}_{\cS\cO_0})$ with the property that
$\fM^{nc}(T,T')\in \cV(\fT^{pl,L}_{\cS\cO_0})$. This gives a non-associative,
non-commutative partial algebra $\cA_{na,nc,L}=(\cV(\fT^{pl,L}_{\cS\cO_0}), \fM^{nc,L})$.
The vector space $\cV(\fF^{pl,L}_{\cS\cO_0})$ can similarly be regarded both as
a quotient of $\cV(\fF^{pl}_{\cS\cO_0})$ under the quotient map $\Pi_L$ or as
a subspace. We can consider on $\cV(\fF^{pl,L}_{\cS\cO_0})$ a coproduct
induced by the coproduct $\Delta$ of $\cV(\fF^{pl}_{\cS\cO_0})$, determined by setting
$$ \Delta_L (T) =\sum_{v\in V_{int}(T): T_v, T/T_v\in \fT^{pl,L}} T_v \otimes (T/T_v)\, . $$
The induced action of $\cA_{na,nc,L}$ on $\cV(\fF^{pl,L}_{\cS\cO_0})$ is given by
$$ \rho^{pl,L}(T)(F)=\Pi_L \circ (\fM^{T,nc}\otimes 1) \circ \Delta_L (F) \, , $$
and satisfies by construction the stated compatibility.
\endproof

\smallskip

We then see that the combination of the procedure described in \S \ref{OrderSec}
for introducing linear ordering of sentences, with the quotient procedure that
eliminates trees that are not realizable as syntactic trees of a specific language,
describes the externalization process in the form of a correspondence, in the
sense outlined in \S \ref{CorrSec} above. 

\smallskip

\begin{defn}\label{externDef}
Externalization is a correspondence given by the span of algebras (or partial algebras) and associated
modules
\begin{equation}\label{exteq}
\xymatrix{ & \cA_{na,nc,L} \otimes \cV(\fF^{pl,L}_{\cS\cO_0}) \ar[r]^{\ \ \ \ \ \rho^{pl,L}} & 
\cV(\fF^{pl,L}_{\cS\cO_0}) \\
\cA_{na,nc} \otimes \cV(\fF^{pl}_{\cS\cO_0}) \ar[r]^{\ \ \ \rho^{pl}} \ar[d]^{\Pi\otimes \Pi}\ar[ur]^{\Pi_L\otimes \Pi_L} & \cV(\fF^{pl}_{\cS\cO_0})\ar[d]^{\Pi} \ar[ur]_{\Pi_L} &  \\ 
 \cA_{na,c} \otimes \cV(\fF_{\cS\cO_0}) \ar[r]^{\ \ \rho} & \cV(\fF_{\cS\cO_0})&  \, .}
\end{equation}
\end{defn}

\smallskip
\subsection{The role of syntactic parameters} \label{ParamSec}

In the Minimalist Model, where the core structure of syntax is
described by the Merge operation of binary set formation, 
{\em syntactic parameters}, which account for syntactic variation
across languages, become part of the externalization structure. 
The notion of syntactic parameters was originally introduced in the
context of the Principles and Parameters model, \cite{ChomskyPP},
\cite{ChoLa}. A recent extensive study of syntactic
parameters can be found in \cite{Roberts}. 

\smallskip

For simplicity, we can assume that syntactic parameters are binary variables.
This may not account for phenomena such as some kind of entailment
relations between parameters, observed for example in \cite{LongGua}, 
but it is still, to a large extent, accurate. 
We can describe the set of syntactic parameters as a subset
$\cP \subset \F_2^N$, where $N$ is a (large) number of binary
variables that record various syntactic features of languages,  
and the locus $\cP \subset \F_2^N$ accounts for the set of
``possible languages" (possible values of parameters that
can be realized by actual human languages, see \cite{Moro}).
The set $\cP$ incorporates all the possible 
relations between parameters. One knows a significant number of
relations is expected, for example through the geometric and topological
data analysis techniques applied to databases of syntactic features, see
for instance \cite{GakMar}, \cite{OrtBerMar}, \cite{PortKarMar},  \cite{ShuMar}.
The exact nature of these relations is not known, but one can hypothesize 
that $\cP$ may be realizable as an algebraic set (or algebraic variety) 
over $\F_2$, embedded in the affine space $\F_2^N$. Regardless of any
specific assumption on the geometry of the set $\cP$, we have that
a language $L$ determines a corresponding point $\pi_L \in \cP$,
which is a vector $\pi_L \in \F_2^N$ that lists as entries the
binary values of the $N$ syntactic parameters for that particular language. 

\smallskip

In the description of externalization proposed in \S \ref{ExtCorrSec},
one expects that syntactic parameters will be involved in
determining both the section $\sigma_L$ of \eqref{sigmaL} and the projection 
$\Pi_L$ of \eqref{quotL}.

\smallskip

Since the first part of externalization, which corresponds to the
section $\sigma_L$ of \eqref{sigmaL}, only depends on syntactic
parameters that govern word order, while the projection 
$\Pi_L$ of \eqref{quotL} depends on all other parameters, 
we can single out a subset of $M<N$ parameters that affect
word-order. We denote by $q: \F_2^N \to \F_2^M$ the corresponding projection
map that only keeps the word-order parameters, and we denote
by $\bar\cP=q(\cP)$ the image under this projection of the locus of
parameters, with $\bar\pi=q(\pi)$, for $\pi\in \cP$. The parameters
$q(\pi_L)_i$, $i=1,\ldots, M$ of a point $q(\pi_L)$ in this space 
$\bar\cP\subset \F_2^M$ cut out a subset of $\fT^{pl}_{\cS\cO_0}$
that consists of those planar structures for trees in $\fT_{\cS\cO_0}$
that are compatible with the word-order properties of the given 
language $L$. These define the range of the section $\sigma_L$,
and similarly for workspaces $\fF^{pl}_{\cS\cO_0}$.

\smallskip

On the other hand, given the set of all planar binary trees and forests
in $\fT^{pl}_{\cS\cO_0}$ and $\fF^{pl}_{\cS\cO_0}$, respectively, the
syntactic parameters specified by the point $\pi_L \in \cP$ have the
effect of selecting which syntactic trees are realizable in the given
language $L$, thus determining the sets $\fT^{pl,L}_{\cS\cO_0}$ and 
$\fF^{pl,L}_{\cS\cO_0}$. We can give the following geometric description
of this procedure, which has the advantage that it allows for the
possible use of tools from algebraic geometry to model more closely
the externalization process. We give below some example of questions
that can be naturally formulated in this mathematical framework. 

\smallskip

As discussed in
the vector spaces
$\cV(\fT^{pl}_{\cS\cO_0})$ and $\cV(\fF^{pl}_{\cS\cO_0})$ are graded
by number of leaves (sentence length), 
\begin{equation}\label{gradingFT}
 \cV(\fT^{pl}_{\cS\cO_0})=\oplus_\ell \cV(\fT^{pl}_{\cS\cO_0})_\ell \ \ \text{ and } \ \ \
\cV(\fF^{pl}_{\cS\cO_0}) =\oplus_\ell \cV(\fF^{pl}_{\cS\cO_0})_\ell\, , 
\end{equation}
with finite dimensional graded pieces.

\smallskip

\begin{prop}\label{paramGr}
Let $\cL$ denote the set of languages $L\in \cL$. Let ${\rm Gr}(d, n)$ 
denote the Grassmannian of $d$-dimensional linear subspaces in 
an $n$-dimensional space. 
The identification of the spaces $\cV(\fT^{pl,L}_{\cS\cO_0})$ and
$\cV(\fF^{pl,L}_{\cS\cO_0})$ by specifying the syntactic parameters
$\pi_L\in \F_2^N$ for a language $L$ is described by a collection of maps
\begin{equation}\label{mapsGr}
E_{i,\ell} : \cP \to {\rm Gr}(d_{\pi_i,\ell}, d_\ell) \, ,
\end{equation}
where for $\pi=(\pi_i)_{i=1}^N \in \cP$, the image
$E_{i,\ell}(\pi) \subset \cV(\fT^{pl}_{\cS\cO_0})_\ell$ is the subspace
spanned by the trees that are compatible with the constraints imposed by
the value of the $i$th syntactic parameter $\pi_i$, with $d_\ell=\dim \cV(\fT^{pl}_{\cS\cO_0})_\ell$.
Thus, a point $\pi \in \cP$ determines
a subspace $E_\ell(\pi)=\cap_i E_{i,\ell}(\pi)$, and similarly for $\cV(\fF^{pl}_{\cS\cO_0})$. 
The assignment $\pi: \cL \to \cP$ of syntactic parameters to languages $L \mapsto \pi_L$
in turn determines $\cV(\fT^{pl,L}_{\cS\cO_0})$ as
\begin{equation}\label{sumEell}
 \cV(\fT^{pl,L}_{\cS\cO_0}) =\oplus_\ell E_{i,\ell}(\pi_L) \, , 
\end{equation} 
and similarly for $\cV(\fF^{pl,L}_{\cS\cO_0})$.
\end{prop}

\proof As in Lemma~\ref{subalgL}, we can view $\cV(\fT^{pl,L}_{\cS\cO_0})$
and $\cV(\fF^{pl,L}_{\cS\cO_0})$ as subspaces (rather than quotient spaces)
of $\cV(\fT^{pl}_{\cS\cO_0})$ and $\cV(\fF^{pl}_{\cS\cO_0})$, respectively. 
Each syntactic parameter $\pi_i$ of a point $\pi=(\pi_i)_{i=1}^N \in \cP\subset 
\F_2^N$ determines a subspace $\cV(\fT^{pl}_{\cS\cO_0})_{\pi_i}\subset  \cV(\fT^{pl}_{\cS\cO_0})$
(respectively, $\cV(\fF^{pl}_{\cS\cO_0})_{\pi_i} \subset \cV(\fF^{pl}_{\cS\cO_0})$, such that,
for $\pi=\pi_L$ for some language $L\in \cL$
\begin{equation}\label{capiParam}
 \bigcap_{i=1}^N \cV(\fT^{pl}_{\cS\cO_0})_{\pi_{L,i}} = \cV(\fT^{pl,L}_{\cS\cO_0}) \, , 
\end{equation} 
and similarly for the $\cV(\fF^{pl}_{\cS\cO_0})_{\pi_{L,i}}$. Given the graded
structure \eqref{gradingFT},
we can consider the procedure \eqref{capiParam} of cutting out the subspaces
$\cV(\fT^{pl,L}_{\cS\cO_0})$ and $\cV(\fF^{pl,L}_{\cS\cO_0})$ step
by step by degrees. For a given $\ell \in \N$, there are integers $c_{\pi_i, \ell}$, $i=1,\ldots, N$
that specify the codimensions of the subspaces
$$ \cV(\fT^{pl}_{\cS\cO_0})_{\pi_i,\ell} \subset \cV(\fT^{pl}_{\cS\cO_0})_\ell\, . $$
If $d_\ell=\dim \cV(\fT^{pl}_{\cS\cO_0})_\ell$ with $d_{\pi_i,\ell}=d_\ell-c_{\pi_i, \ell}$ the dimensions, 
we then have maps
$$ E_{i,\ell}: \cP \to {\rm Gr}(d_{\pi_i,\ell}, d_\ell) $$
to the Grassmannian of $d_{\pi_i,\ell}$-dimensional subspaces inside
the $d_\ell$-dimensional space $\cV(\fT^{pl}_{\cS\cO_0})_\ell$,  so that 
$$ E_{i,\ell}(\pi_L)=\cV(\fT^{pl}_{\cS\cO_0})_{\pi_{L,i}}  \in {\rm Gr}(d_{\pi_{L,i},\ell}, d_\ell)\, ,$$
and similarly for $\cV(\fF^{pl}_{\cS\cO_0})_\ell$. Similarly, if
$d_{L,\ell}=\dim \cV(\fT^{pl,L}_{\cS\cO_0})_\ell$ is the dimension of the resulting
intersection (which need not be transversal due to relations between
syntactic parameters), the vector $\pi_L\in \F_2^N$ of parameters for the
language $L$ determines a map 
$$ E_{\ell}\circ \pi: \cL \to \bigcup_d {\rm Gr}(d, d_\ell) \ \ \ \ \ 
 E_{\ell}(\pi_L)=\cV(\fT^{pl,L}_{\cS\cO_0})_\ell \in {\rm Gr}(d_{L,\ell}, d_\ell)\, , $$
 with $\cV(\fT^{pl,L}_{\cS\cO_0})=\oplus_\ell E_{\ell}(\pi_L)$.
\endproof

\smallskip

There are natural geometric questions that this viewpoint suggests.
For instance, when comparing the syntax of different languages
$L,L'\in \cL$, one can consider the resulting comparison between
the systems of subspaces $\cV(\fT^{pl,L}_{\cS\cO_0})=\oplus_\ell E_{\ell}(\pi_L)$ 
and $\cV(\fT^{pl,L'}_{\cS\cO_0})=\oplus_\ell E_{\ell}(\pi_{L'})$. 
Syntactic proximity can be viewed in terms of the geometric position of
these subspaces. For example, mathematically a special case of
pairs $E,E'$ of infinite dimensional subspaces inside an infinite
dimensional space $\cV$ is given by the Fredholm pairs, where
the intersection $E\cap E'$ is finite dimensional and the
span of the union $E\cup E'$ has finite codimension. These would
represent the situation of maximal differentiation. 
Moreover, there are models of semantics based on the geometry
of Grassmannians, \cite{ManMar}, and one can consider in this
context the possibility of algebro-geometric models of a syntactic-semantic
inferface. 

\section{Merge and fundamental combinatorial recursions in physics}\label{PhysSec}

In classical physics, a ``least action principle" governs the solutions of equations of
motion of physical systems, in the form of minimization (or stationarity) of the action
functional, namely a minimization with respect to energy.  The equations of motion
are then expressed as the Euler--Lagrange equations that describe the stationarity
of the action functional under infinitesimal variation.
In quantum physics, and more precisely in quantum field theory, the classical
equations of motion become equations in the quantum correlation functions of
the fields (see \cite{PeS}, \S 9.6). More precisely, the Euler--Lagrange equations
are satisfied by the Green functions of the quantum field theory, up to terms
that reflect the noncommutativity of field operators. The resulting quantum equations 
of motion are known in physics as Dyson--Schwinger equations. They represent
the optimization process of the least action principle, implemented at the quantum level.

\smallskip

Quantum physics, in the form of perturbative quantum field theory, is governed
by a combinatorial generative process that determines the terms of the perturbative 
expansion. The combinatorial objects involved are the Feynman graphs of the
theory, and the generative process can be described either by formal languages
(in the form of graph grammars, see \cite{MaPo}) or in a more efficient way in
terms of Hopf algebras (the Connes--Kreimer Hopf algebras of Feynman graphs
and of rooted trees, see \cite{CoKr}, \cite{CoMa}). 

\smallskip

These two different descriptions
of the generative process that produces the Feynman graphs of quantum field
theory can be compared to what happens with older formulations of the
Minimalist Model in generative linguistics, where one can give both a formal 
languages description (see \cite{VSW}) and a description in terms of (internal/external) 
Merge operators, where the latter is computationally significantly more efficient 
(see \cite{Berwick}). 

\smallskip

We discuss in a separate companion paper \cite{MBC} how to compare older
versions of the Minimalist Model to the new version of \cite{Chomsky17}, \cite{Chomsky19} 
that we analyzed in this paper, at the level of the Hopf algebra structure, and how one
sees in those terms the advantage of the more recent formulation. 

\smallskip

Here the main point we want to stress is that, in the setting of quantum physics, the best 
description of the
generative process of the hierarchy of Feynman graphs organized by increasing loop
number in the asymptotic expansion is also determined by a Hopf algebra. There
are two main advantages of this algebraic formalism in physics:
\begin{enumerate}
\item The algebraic structure governs the construction of the quantum solutions 
of the equations of motion, through the Dyson--Schwinger equations recalled above,
so that solutions can be constructed through a combinatorial recursive procedure.
\item The Hopf algebra formalism also transparently explains the
renormalization process in physics (namely the elimination of infinities, that is, the 
consistent extraction of finite (meaningful) values from divergent Feynman integrals). 
\end{enumerate}

\smallskip

We will discuss more in detail here the role of the algebraic
formalism in quantum field theory in the recursive construction of
solutions to the Dyson--Schwinger equations, as this is the aspect that
is more closely related to the properties of Merge that we discussed in
the previous sections. We will only sketch briefly in \S \ref{RenormSec}
the possible relevance of the algebraic formulation of renormalization
to linguistics models, as we plan to return to discuss that in a separate paper.

\smallskip
\subsection{The recursive construction of Dyson--Schwinger equations}

In the physics of renormalization in quantum field theory, the generative
process for the hierarchical structure of Feynman graphs is described
equivalently by the Connes--Kreimer Hopf algebra of Feynman graphs
mentioned above \cite{CoKr}, or by a Hopf algebra of planar rooted trees
(not necessarily binary), where the tree structure describes the way in which
subgraphs are nested inside Feynman graphs (see \cite{Foissy2}, \cite{Kr}).  
When formulated in terms of the Hopf algebra of trees, one can obtain a 
recursive construction of the solutions of the equations of motion of the 
quantum system, the Dyson--Schwinger equations, in terms of the 
combinatorics of trees, see \cite{BergKr}, \cite{Foissy}, \cite{Yeats}.

\smallskip

This happens in the following way, as we outlined briefly in  \S \ref{ActMergeSec} and \S \ref{CoreSec}.
The Hopf algebra $\cH$ of planar rooted trees and forests has 
product given by disjoint union and coproduct given by
$$ \Delta(T)=\sum_C \pi_C(T)\otimes \rho_C(T) \, , $$
where the left-hand-side $\pi_C(T)$ of the coproduct is a forest obtained
by cutting subtrees of $T$ using an ``admissible cut" (not two cut legs on the
same path from root to leaves) and the right-hand-side $\rho_C(T)$ is
the tree that remains attached to the root when the cut is performed. Note
that this is a form of the coproduct that we also used in \eqref{coprodT}.

\smallskip

One defines an operators $B^+ : \cH \to \cH$, 
as in Definition~\ref{Bplusop}. Namely, $B^+$ acts on a forest
$T_1 \sqcup \cdots \sqcup T_m$  by creating a new rooted tree $T$
where all the roots $v_{r_1}, \ldots, v_{r_m}$ of the trees $T_1, \ldots, T_m$ 
are attached to a single new root vertex,
$$ B(T_1\cdot \cdots \cdot T_m) = T  = \Tree[ $T_1$ $T_2$ $\cdots$ $T_n$ ] \, . $$
As we observed in \S \ref{ActMergeSec} and \S \ref{CoreSec}, this 
has exactly the structure of a Merge operator (though not
necessarily binary, as it can take an arbitrary number of input trees). 
The operator $B^+$ satisfies the identity
\begin{equation}\label{cocycle}
 \Delta( B^+(X)) = B^+(X) \otimes 1 + (Id \otimes B^+)\circ \Delta (X), 
\end{equation} 
for all $X\in \cH$. This identity is the Hochschild $1$-cocycle condition (see \cite{CoKr}, \cite{BergKr}, \cite{Foissy2}). 

\smallskip

The combinatorial Dyson--Schwinger equation then takes the form of a fixed point
equation
\begin{equation}\label{DSeq}
X = B^+ (P(X)) ,
\end{equation}
where $X=\sum_{k\geq 1} x_k$ is a formal series of elements $x_k \in \cH_k$ in
the graded pieces of the Hopf algebra, and $P(t)=\sum_{k\geq 0} a_k t^k$ with $a_0=1$
is a formal power series (or polynomial). The simplest and most fundamental such
equation is the case where $P$ is quadratic, $P(X)=X^2$, which is the form that we have
encountered in \eqref{DStreesMerge}, which governs the generative process of the 
core structure of Merge, discussed in \S \ref{CoreSec}. 
The equation \eqref{DSeq} has a unique solution $X=\sum_{k\geq 1} x_k$ (\cite{BergKr}, \cite{Foissy2}) 
that can be written in the recursive form
\begin{equation}\label{DSrecsol}
 x_{n+1} = \sum_{k=1}^n \sum_{j_1+\cdots+j_k=n} a_k B^+(x_{j_1} \cdots x_{j_k}), 
\end{equation} 
with initial step $x_1 = B^+(1)$. 
It is shown in \cite{BergKr}, \cite{Foissy2} that the cocycle
property \eqref{cocycle} of the operator $B$ is required to ensure that the 
coordinates $x_n$ of the solution of a Dyson--Schwinger equation determine
a Hopf subalgebra, though the construction of the solution \eqref{DSrecsol} itself
does not require the cocycle condition \eqref{cocycle}. In the case
of the linguistic Merge the basic combinatorial structure is the same, with the 
recursion \eqref{DSrecsol}
corresponding to the core generative process of Merge, as we described in \S \ref{CoreSec}. 

\smallskip

For a short overview of how this kind of combinatorial Dyson--Schwinger equations recover the
physical equations of motion in quantum field theory, see \cite{Wein}. For a more
detailed treatment of combinatorial Dyson--Schwinger equations see \cite{Yeats}.
A discussion of the use of Dyson--Schwinger equation in the context of the theory of
computation is given in \cite{DelMar}, following the approach to Renormalization and
Computation developed in \cite{Man1}, \cite{Man2}.

\smallskip

As we mentioned at the beginning of this section, the classical Euler--Lagrange equations of motion
express an optimality process given by a least action principle, and the quantum equations of motion
given by the recursively solved Dyson--Schwinger equations, reflect this form of optimization in the
quantum setting. In this sense the core computational structure of syntax defined by the Merge
operator can be seen as being optimal and most fundamental, as it reflects the structure of the
physical Dyson--Schwinger equation (for the appropriate Hopf algebra) and for the most
basic (quadratic) form of the recursion. 

\smallskip

One can ask whether there are any other characterizations of the syntactic Merge
by optimality. The optimization is usually done with respect to some real-valued
cost functional (energy/action in the case of physical systems). There are also other
ways of thinking of optimization that do not require an evaluation through a function
with values in real numbers. For example, it is possible to formulate optimization
processes in a purely categorical framework (see for instance \cite{Mar2}), and
an optimality property for the syntactic Merge may similarly take some more
abstract categorical form. On the other hand it is also possible to consider 
minimization conditions with respect to other types of ``action functional" that
replace energy in the case of computational systems. For example, it is argued
in \cite{Man3} that complexity provides a suitable replacement for energy in
the context of the theory of computation. We leave these questions to further
investigation.

\smallskip
\subsection{Algebraic renormalization and its relevance to linguistics models}\label{RenormSec}

The second aspect we mentioned above of the Hopf algebra formalism in
physics is related to the mathematical treatment of renormalization in physics.
The renormalization process in quantum field theory is a fundamental
process that allows for the evaluation of Feynman integrals through
a subtraction of infinities that is compatible with the hierarchical generative
process of graphs (that is, with the subdivergences nested inside larger
graphs). More precisely, one wants to extract a ``meaningful" (finite) part
out of the calculation of a priori divergent Feynman integrals, in such a
way that this extraction of the ``meaningful part" can be performed compatibly
with what already assigned to subgraphs inside of larger graphs (that is,
renormalization of sub-divergences, in physics terms). 

\smallskip

An analogy with the linguistics setting immediately comes to mind, where
one literally talks about assignment of meaning (semantics) to syntactic
parsing trees of sentences, discarding possibilities that are ruled out by
semantics. One similarly encounters the requirement that such assignment
be done consistently across subtrees. For the moment this is purely an
analogy, but one can see that this can be enriched with actual precise
mathematical content, by considering an intermediate step between
physics and linguistics, which is the extension of the formalism of
renormalization and Hopf algebra from physics to the theory of computation,
developed by Manin in  \cite{Man1}, \cite{Man2}. In this setting, one deals,
as in physics, with a problem of subtraction of infinities, which arise not
from divergence of Feynman integrals but from non-computability (in the
form of divergence of computational time in the halting problem). A
procedure of extraction of computable ``subfunctions" from 
non-decidable problems is organized in terms of a Hopf algebra of flow charts (of algorithms)
replacing the Hopf algebra of Feynman graphs. The role of Dyson--Schwinger
relations in this context was analyzed in \cite{DelMar}. 

\smallskip

More precisely, 
in the mathematical formulation of renormalization (see \cite{CoKr}, \cite{CoMa}),
the (regularized) Feynman rules are described as a morphism of commutative
algebras $\Phi: \cH \to \cR$, where $\cH$ is the Hopf algebra of Feynman graphs
and $\cR$ is an algebra of functions where regularization
of Feynman integrals takes place, such as Laurent series, with the structure
of Rota--Baxter (RB) algebra of weight $-1$. The  RB structure captures the properties
of the subtraction of the polar part of the Laurent series and more general it
models a good process of subtraction of a divergent part. The coalgebra and
antipode on $\cH$, together with the Rota--Baxter operator on $\cR$, determine
a Birkhoff factorization of $\Phi$ into a part $\Phi_-$ that carries all the divergences
and a part $\Phi_+$ that gives the finite renormalized Feynman amplitudes (the actual
physical quantities). This operation carries within itself the consistent assignment
of the meaningful convergent part across subgraphs, with the coproduct of the
Hopf algebra organizing the subgraphs. 

\smallskip

We will not discuss any further in the present paper the linguistic analog for
this Birkhoff factorization, although we can mention a possible interesting direction
of investigation. A version of Birkhoff factorization taking place in semirings, with
applications to the theory of computation was developed in \cite{Mar}, \cite{MaTe}. 
There are models of syntactic parsing and of semantics that are based on the same
mathematical structure of semirings (see for instance \cite{Good}, \cite{Roak}).
One can expect that it should be possible to construct a mathematical
model of a syntactic-semantic interface that resembles the physical model of 
renormalization, with extraction of semantic meaning replacing the extraction
of renormalized physical values. We will consider this problem elsewhere,
as a separate paper.

\bigskip
\subsection*{Acknowledgments}
We thank Riny Huijbregts and Sandiway Fong for providing extensive
comments and very helpful feedback on a draft of this paper. 
The first author acknowledges support from NSF grant DMS-2104330 and 
FQXi grants FQXi-RFP-1 804 and FQXi-RFP-CPW-2014, SVCF grant 2020-224047, 
and support from the Center for Evolutionary Science at Caltech.

\end{document}